\documentclass[twocolumn,natbib]{my_arXiv}
\usepackage{graphicx}
\usepackage{times}
\usepackage{epsfig}
\usepackage{amsmath}
\usepackage{amssymb}

\usepackage{natbib}

\usepackage[shortcuts]{extdash}
\usepackage{enumerate}
\usepackage{picins}
\usepackage{multirow}
\usepackage[USenglish]{babel}
\usepackage[ruled,vlined]{algorithm2e} 
\usepackage{float}

\usepackage[usenames]{color} 
\definecolor{Red}{rgb}{1.0, 0, 0} 
\definecolor{coarse}{rgb}{0.753, 0.314, 0.302} 
\definecolor{fine}{rgb}{0.310, 0.506, 0.741}   
\definecolor{ours}{rgb}{.8, 0, 0} 
\definecolor{citecl}{rgb}{0, 0, .8} 

\usepackage{bbm}
\usepackage{bbold}

\usepackage[hyphens]{url}

\usepackage{floatflt}

\newcommand{\abs}[1]{\mbox{$\left|#1\right|$}}

\newcommand{\VV}{\mbox{$\mathcal{V}$}}
\newcommand{\EE}{\mbox{$\mathcal{E}$}}

\newcommand{\deff}{\stackrel{\tiny{\mbox{def}}}{=}}

\newlength{\origtabcolsep}
\begin{document}

\title{A Unified Multiscale Framework for Discrete Energy Minimization}


\author{Shai Bagon         \and
        Meirav Galun      
}

\institute{S. Bagon \at
              Weizmann Inst. of Science \\
              Tel.: +972-8-9344268\\
              \email{shai.bagon@weizmann.ac.il}           
           \and
           M. Galun \at
              Weizmann Inst. of Science \\
              Tel.: +972-8-9342141\\
              \email{meirav.galun@weizmann.ac.il}           
}

\date{}

\maketitle

\begin{abstract}
Discrete energy minimization is a ubiquitous task in computer vision, yet is NP\-/hard in most cases.
In this work we propose a multiscale framework for coping with the NP\-/hardness of discrete optimization.
Our approach utilizes algebraic multiscale principles to efficiently explore the discrete solution space,
yielding improved results on challenging, non\-/submodular energies for which current methods provide unsatisfactory approximations.
In contrast to popular multiscale methods in computer vision, that builds an {\em image pyramid},
our framework acts directly on the energy to construct an {\em energy pyramid}.
Deriving a multiscale scheme from the energy itself makes our framework application independent and widely applicable.
Our framework gives rise to two complementary energy coarsening strategies:
one in which coarser scales involve fewer variables,
and a more revolutionary one in which the coarser scales involve fewer discrete labels.
We empirically evaluated our unified framework on a variety of both non\-/submodular and submodular energies, including energies from Middlebury benchmark.
\end{abstract}

\section{Introduction}

\begin{figure}[t]

\centering
\framebox[\linewidth][c]{
\parbox{\linewidth}{
\centering
\includegraphics[width=\linewidth]{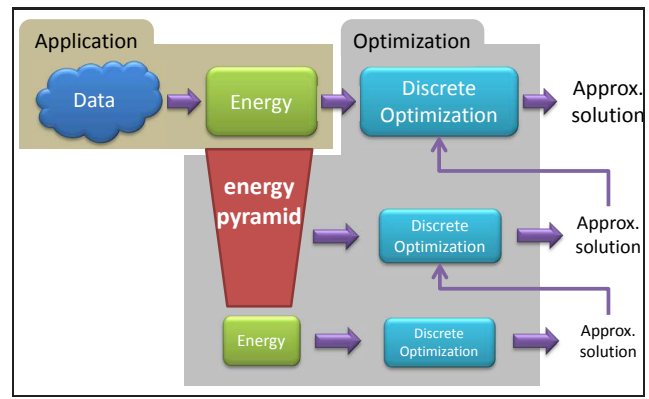}
}}
\vspace*{1mm}
\caption{
{\bf A Unified multiscale framework:}
{\em
We derive multiscale representation of the energy itself = energy pyramid.
Our multiscale framework is unified in the sense that different problems with different energies share the same multiscale scheme, making our framework widely applicable and general.
}}
\label{fig:multiscale-schemes}
\end{figure}


Discrete energy minimization is ubiquitous in computer vision, and spans a variety of problems.
These energies can be grossly divided into two classes: submodular and non\-/submodular energies.
Submodular energies are characterized by ``smoothness" encouraging pairwise (or higher order) terms.
Apart from the binary case, minimizing these energies is known to be NP-hard.
Despite this theoretical hardness, such submodular energies, which naturally reflect a ``piecewise constant" prior,  gained popularity and became very common in computer vision applications, such as denoising, stereo and multi-label segmentation (e.g., \cite{Szeliski2008}).
For this reason most of the efforts of the vision community regarding discrete optimization focused on developing approximate optimization methods for these submodular energies, yielding quite successful algorithms.
Recently, more challenging, non\-/submodular energies started to gain popularity.
These energies are characterized by a combination of ``smooth" and ``non-smooth" encouraging pairwise terms.
The correlation-clustering functional, recently applied to segmentation, co-segmentation and clustering (e.g., \cite{Glasner2011,Bagon2012}), is an example for such non\-/submodular energy.
Moreover, non\-/submodular energies may appear when the parameters of the energy are automatically learned (e.g., \cite{Nowozin2011}).
Since such non\-/submodular energies are only recently explored, their optimization receives less attention, and consequently, the existing optimization methods provide approximations that may be quite unsatisfactory.
In practice, it is generally considered a more challenging task to optimize  non\-/submodular energies.

\begin{floatingfigure}[right]{.45\linewidth}
\hspace*{-.05\linewidth}
\includegraphics[width=.43\linewidth]{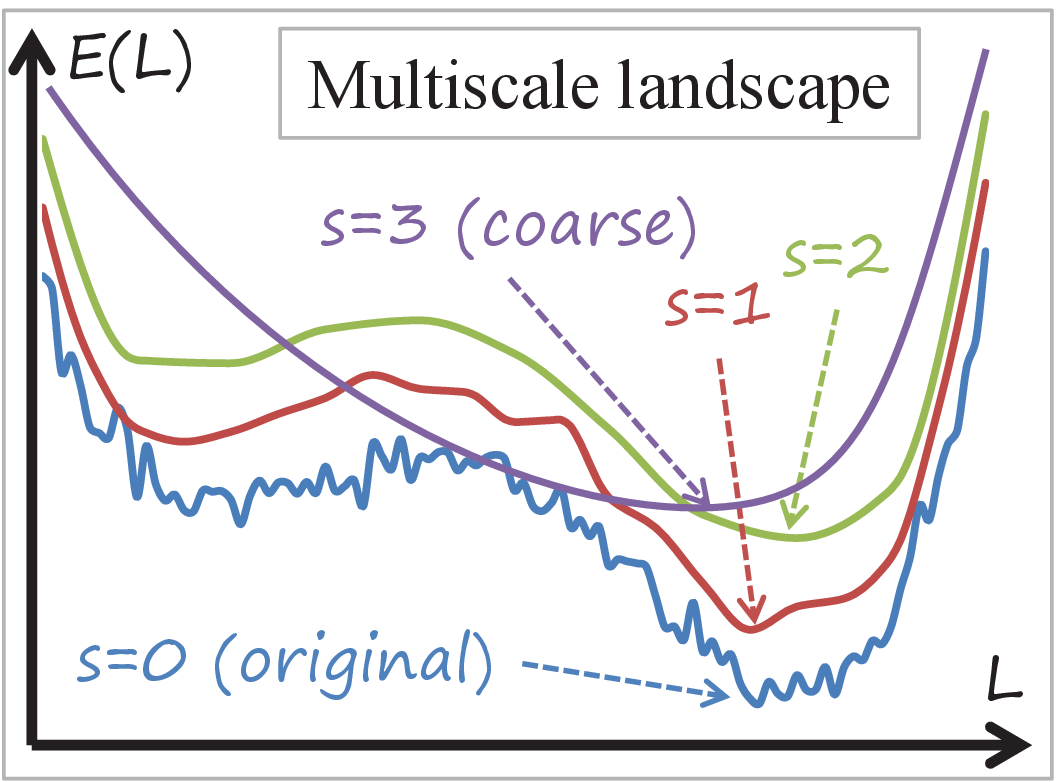}
\end{floatingfigure}
But what makes discrete energy minimization such a challenging endeavor?
The fact that this minimization
implies an exploration of an exponentially large search space.
One way to alleviate this difficulty is to use multiscale search.
The illustration on the right shows
a toy ``energy" $E(L)$ at different scales of detail.
Considering only the original scale ($s=0$), it is very difficult to suggest an effective exploration (optimization) method.
However, when looking at coarser scales ($s=1,\ldots,3$) of the energy an interesting phenomenon is revealed.
At the coarsest scale ($s=3$) the large basins of attraction emerge, but with very low accuracy.
As the scales become finer ($s=2,\ldots,0$), one ``loses sight" of the large basins, but may now ``sense" more local properties with higher accuracy.
We term this well known phenomenon as the {\em multiscale landscape} of the energy.
This multiscale landscape phenomenon encourages coarse\-/to\-/fine exploration strategies:
starting with the large basins that are apparent at coarse scales,
and then gradually and locally refining the search at finer scales.

For more than three decades the vision community focuses on the multiscale pyramid of {\em images} (e.g., \cite{Lucas1981,Burt1983}).
There is almost no experience and no methods that
apply a multiscale scheme {\em directly to discrete energies}.

Another domain in which multiscale methods are common practice is numerical PDE solvers.
Early works in that domain applied {\em geometric} coarsening (geometric multigrid), which is the analogue of the classical image pyramid.
A solution for a PDE was then obtained by applying a single\-/scale solver at each scale (relaxation).
This geometric multigrid paradigm suggested a very simple construction of a regular pyramid at the cost of very careful design of single\-/scale solvers, tailoring them for each problem separately.
A breakthrough for the PDE community was the development of {\em algebraic} multigrid (AMG) of \cite{Brandt1986}.
The {\em algebraic} multigrid approach suggests to derive the pyramid directly from the underlying problem, resulting with irregular data\-/driven pyramid.
This way, local and general solvers (e.g., Gauss-Seidel relaxation) can be incorporated  into the algebraic pyramid yielding improved and robust solutions (\cite{Stuben1999}).

In this paper we present a novel unified discrete multiscale optimization scheme that acts {\em directly} on the energy
(Fig.~\ref{fig:multiscale-schemes}).
Our multiscale framework is unified in the sense that it is application independent: different problems with different energies {\em share the same} multiscale scheme, making our framework widely applicable and general.
More importantly, our multiscale method efficiently explores the discrete solution space through an irregular {\em multiscale energy pyramid}, constructed by {\em energy\-/aware} coarse-to-fine interpolation.
In a sense, our method may be considered as the discrete analogue of AMG:
Instead of focusing attention on complicated optimization schemes,
our framework exposes the multiscale landscape of the energy through energy\-/aware construction of the pyramid.
This way even simple and local optimization methods can be incorporated into our pyramid yielding improved and robust approximations.
In practice, we apply our multiscale optimization method to a large set of challenging problems, including submodular and non\-/submodular, and achieve comparable or lower energy values, than those obtained by the state\-/of\-/the\-/art methods.

\noindent This work makes several contributions:\*
\begin{enumerate}[(i)]
\item A novel unified multiscale framework for discrete optimization:
A wide variety of optimization problems, including segmentation, stereo, denoising, correlation\-/clustering, and others share the same multiscale framework.

\item Any multiscale scheme requires a single\-/scale optimization method to refine the search at each scale.
Our framework is also unified in the sense that it is not restricted to any specific optimization method.

\item Energy\-/aware coarsening scheme.
Variable aggregation takes into account the underlying structure of the energy itself, thus efficiently and directly exposes its multiscale landscape.

\item Provide discrete analogue to AMG.
Incorporating even simple and local optimization methods into out energy\-/aware pyramid yields good approximations.

\item Coarsening the labels.
Our formulation allows for {\em variable} coarsening as well as for {\em label} coarsening.

\item Optimizing hard non\-/submodular energies.
We achieve significantly lower energy assignments on diverse computer vision energies, including challenging non\-/submodular examples.

\end{enumerate}

\subsection{Related work}

Algorithms for discrete energy minimization can work in the primal space or the dual space.
Primal methods act on the discrete variables in the label space to minimize the energy (e.g., \cite{Besag1986,Veksler2002,Rother2007}).
Dual methods formulate a dual problem to the energy and maximize a lower bound to the sought energy (e.g., \cite{Kolmogorov2006}).
Dual methods are recently considered more favorable since they do not only provide an approximate solution, but also provide a lower bound on how far this solution is from the global optimum.
Furthermore, if a labeling is found with energy equals to the lower bound a certificate is provided that the global optimum was found.
For the submodular energies it was shown (by \cite{Szeliski2008}) that dual methods tend to  provide better approximations with very tight lower bounds.
However, using several classes of non\-/submodular energies, we empirically demonstrate that when it comes to challenging non\-/submodular energies, primal methods tend to provide better approximations than dual methods,
since in these cases the lower bound is no longer tight (\cite{Werner2010}).

Our multiscale framework constructs a multiscale energy pyramid in terms of the primal space.
We achieve comparable performance when applied to submodular problems and superior performance when applied to non\-/submodular problems, while comparing it to the state-of-the-art methods (primal and dual).

There are very few works that apply multiscale schemes directly to the discrete energy.
A prominent example for this approach was suggested by \cite{Felzenszwalb2006}; it provides a coarse\-/to\-/fine belief propagation scheme restricted to regular diadic pyramid.
A more recent work is that of \cite{Komodakis2010} that provides an algebraic multigrid formulation for discrete optimization in the dual space.
However, despite his general formulation \citeauthor{Komodakis2010} only provides examples using regular diadic grids of  submodular energies.

The work of \cite{Kim2011} proposes a two-scale scheme mainly aimed at improving run-time of the optimization process.
Their proposed coarsening strategies can be interpreted as special cases of our unified framework.
We analyze their underlying assumptions (Sec.~\ref{sec:local-correlations}), and suggest better methods for efficient exploration of the multiscale landscape of the energy.

The complexity of the optimization algorithms is affected by the number of discrete labels, as well as the number of variables.
Existing optimization algorithms starts to fall behind when facing energies with large label space.
\cite{Lempitsky2007} proposed a method to exploit known properties of the metric between the labels to allow for faster minimization of energies with large number of {\em labels}.
However, their method is restricted to energies with clear and known label metrics and requires training.
In contrast, our framework addresses this issue via a principled scheme that builds an energy pyramid with {\em decreasing number of labels} without prior training and with fewer assumptions on the labels interactions.

\section{Multiscale Energy Pyramid}
\label{sec:unified}

We consider discrete pair-wise minimization problems, defined over a (weighted) graph $\left(\VV, \EE\right)$, of the form:
\begin{eqnarray}
E\left(L\right)&=&\sum_{i\in\VV} \varphi_i\left(l_i\right) + \sum_{\left(i,j\right)\in\EE} w_{ij}\cdot \varphi\left(l_i,l_j\right) \label{eq:GenEng}
\end{eqnarray}
where $\VV$ is the set of variables, $\EE$ is the set of edges, and the solution is discrete: $L\in\left\{1,\ldots,l\right\}^n$, with $n$ variables taking $l$ possible labels.
Many problems in computer vision are cast in the form of~(\ref{eq:GenEng}) (see \cite{Szeliski2008}).
Furthermore, we do not restrict the energy to be submodular, and our framework is
also applicable to more challenging non\-/submodular energies.

Our aim is to build an effective energy pyramid with a decreasing number of degrees of freedom.
The key component in constructing such a pyramid is the interpolation method.
The interpolation maps solutions between levels of the pyramid,
and determines the original energy approximation with fewer degrees of freedom.
We propose a novel principled energy aware interpolation method such that
the resulting energy pyramid efficiently exposes the multiscale landscape of the energy making low energy assignments apparent at coarse levels.

Practically, it is counter intuitive to directly interpolate discrete label values,
since they usually have only semantic interpretation.
Therefore, we substitute an assignment $L$
by an equivalent binary matrix representation $U\in\left\{0,1\right\}^{n\times l}$.
The rows of $U$ correspond to the variables, and the columns corresponds to labels:
$U_{i,\alpha}=1$ iff variable $i$ is labeled ``$\alpha$" ($l_i=\alpha$).
This representation allows us to interpolate discrete solutions, as will be shown in the subsequent sections.

Expressing the energy (\ref{eq:GenEng}) using $U$ yields a relaxed quadratic representation (\cite{Anand2000}).
This algebraic representation forms the basis for our principled multiscale framework derivation:
\begin{eqnarray}
E\left(U\right)&=&Tr\left(DU^T+WUVU^T\right) \label{eq:EngU} \\
      & \mbox{s.t.} & U\in\left\{0,1\right\}^{n\times l},\ \sum_{\alpha=1}^l U_{i\alpha}=1 \label{eq:const-U}
\end{eqnarray}
where $W=\left\{w_{ij}\right\}$, $D\in\mathbb{R}^{n\times l}$ s.t. $D_{i,\alpha}\deff \varphi_i(\alpha)$, and $V\in\mathbb{R}^{l\times l}$ s.t. $V_{\alpha,\beta}\deff \varphi\left(\alpha,\beta\right)$, $\alpha,\beta\in\left\{1,\ldots,l\right\}$.

An energy over $n$ variables with $l$ labels is now parameterized by $\left(n, l , D, W, V\right)$.

We first describe the energy pyramid construction for a general interpolation matrix $P$,
and defer the detailed description of our novel interpolation to Sec.~\ref{sec:matrix-P}.

\subsubsection*{Energy coarsening by variables}

Let $\left(n^f, l, D^f, W^f, V\right)$ be the fine scale energy.
We wish to generate a coarser representation $\left(n^c, l, D^c, W^c, V\right)$ with $n^c<n^f$.
This representation approximates $E\left(U^f\right)$ using fewer {\em variables}: $U^c$ with only $n^c$ rows.

An interpolation matrix  $P\in\left[0,1\right]^{{n^f}\times{n^c}}$ s.t. $\sum_jP_{ij}=1$ $\forall i$, maps coarse assignment $U^c$ to fine assignment $PU^c$.
For any fine assignment that can be approximated by a coarse assignment $U^c$, i.e.,
\begin{eqnarray}
U^f & \approx & PU^c \label{eq:interp}
\end{eqnarray}

Plugging (\ref{eq:interp}) into~(\ref{eq:EngU}):
\begin{eqnarray}
E\left(U^f\right) & = & Tr\left(D^f{U^f}^T+W^fU^fV{U^f}^T\right) \nonumber \\
& \approx & Tr\left(D^f{U^c}^TP^T+W^fPU^cV{U^c}^TP^T\right) \nonumber \\
& = & Tr\Big(\underbrace{\left(P^TD^f\right)}_{\mbox{\normalsize $ \deff D^c$ }}{U^c}^T + \underbrace{\left(P^TW^fP\right)}_{\mbox{\normalsize $\deff W^c$}}U^cV{U^c}^T\Big) \nonumber \\
& = & Tr\left(D^c{U^c}^T+W^cU^cV{U^c}^T\right) \nonumber \\
& = & E\left(U^c\right)  \label{eq:EngC}
\end{eqnarray}
We have generated a coarse energy $E\left(U^c\right)$
parameterized by $\left(n^c, l, D^c, W^c, V\right)$ that approximates the fine energy $E(U^f)$.
This coarse energy is {\em of the same form} as the original energy allowing us to apply the coarsening procedure recursively to construct an energy pyramid.

\subsubsection*{Energy coarsening by labels}
So far we have explored the reduction of the number of degrees of freedom by reducing the number of {\em variables}.
However, we may just as well look at the problem from a different perspective: reducing the search space by decreasing the number of {\em labels} from $l_f$ to $l_c$ ($l_c<l_f$).
It is a well known fact that optimization algorithms 
suffer from significant degradation in performance as the number of {\em labels} increases (\cite{Bleyer2010}).
Here we propose a novel principled and general framework for reducing the number of labels at each scale.

Let $\left(n, l^f, D^{\hat{f}}, W, V^{\hat{f}}\right)$ be the fine scale energy.
Looking at a  different interpolation matrix $\hat{P}\in\left[0,1\right]^{\mbox{$l^f\times l^c$}}$,
we  interpolate a coarse solution by $U^{\hat{f}} \leftarrow U^{\hat{c}}\hat{P}^T$.
This time the interpolation matrix $\hat{P}$ acts on the {\em labels}, i.e., the {\em columns} of $U$.
The coarse labeling matrix $U^{\hat{c}}$ has the same number of rows (variables), but fewer columns (labels).
We use $\hat{\Box}$
notation to emphasize that the coarsening here affects the labels rather than the variables.

Coarsening the labels yields:
\begin{equation}
E\left(U^{\hat{c}}\right) = Tr\left( \left(D^{\hat{f}}\hat{P}\right)\mbox{$U^{\hat{c}}$}^T + WU^{\hat{c}} \left(\hat{P}^TV^{\hat{f}}\hat{P}\right)\mbox{$U^{\hat{c}}$}^T\right)
\label{eq:EngC-V}
\end{equation}
Again, we end up with the same type of energy, but this time it is defined over a smaller number of discrete labels:
$\left(n, l^c, D^{\hat{c}}, W, V^{\hat{c}}\right)$,
where $D^{\hat{c}} \deff D^{\hat{f}}\hat{P}$ and $V^{\hat{c}} \deff \hat{P}^T V^{\hat{f}} \hat{P}$.

\

The main theoretical contribution of this work is encapsulated in the
multiscale ``trick" of equations~(\ref{eq:EngC}) and~(\ref{eq:EngC-V}).
Formulating the interpolation as a linear operator ($P$) and plugging it in the quadratic energy representation~(\ref{eq:const-U})  provides a principled algebraic representation for our multiscale framework.
Our direct formulation is in contrast to the ``ad-hoc" representation of \cite{Felzenszwalb2006,Kim2011}, and \cite{Komodakis2010}.
Our scheme moves the multiscale completely to the optimization side and makes it independent of any specific application.
We can practically approach now a wide and diverse family of energies using {\em the same} multiscale implementation.

The effectiveness of the multiscale approximation of~(\ref{eq:EngC}) and~(\ref{eq:EngC-V}) heavily depends on the interpolation matrix $P$ ($\hat{P}$ resp.).
Poorly constructed interpolation matrices will fail to expose the multiscale landscape of the functional.
In the subsequent section we describe our principled energy\-/aware method for computing it.


\section{Energy-aware Interpolation}
\label{sec:matrix-P}

\begin{figure}
\centering
\parpic[r][r]{\includegraphics[width=.5\linewidth]{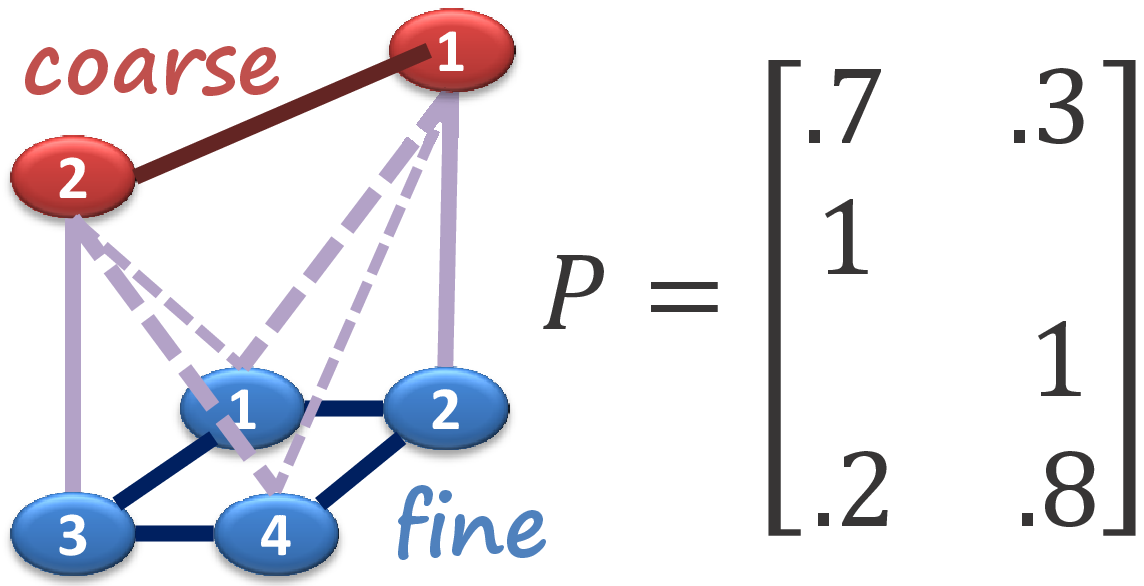}}
\caption{
{\bf Interpolation as soft variable aggregation:}
{\em
{\color{fine} fine} variables {\color{fine}1, 2, 3} and {\color{fine}4} are  softly aggregated into
{\color{coarse}coarse} variables {\color{coarse}1} and {\color{coarse}2}.
For example,  {\color{fine}fine} variable {\color{fine}1} is a convex combination of $.7$ of {\color{coarse}1} and $.3$ of {\color{coarse}2}.
Hard aggregation is a special case where $P$ is a binary matrix.
In that case each fine variable is influenced by exactly one coarse variable.}
}
\label{fig:multiscale}
\end{figure}

In this section we use terms and notations for variable coarsening ($P$),
however the motivation and methods are applicable for label coarsening ($\hat{P}$) as well due to the similar algebraic structure of~(\ref{eq:EngC}) and~(\ref{eq:EngC-V}).

Our energy pyramid approximates the original energy using a decreasing number of degrees of freedom,
thus excluding some solutions from the original search space at coarser scales.
Which solutions are excluded is determined by the interpolation matrix $P$.
{\bf A desired interpolation does not exclude low energy assignments at coarse levels}.

The matrix $P$ can be interpreted as an operator that aggregates fine-scale variables into coarse ones (Fig.~\ref{fig:multiscale}).
Aggregating fine variables $i$ and $j$ into a coarser one excludes from the search space all assignments for which $l_i\ne l_j$.
This aggregation is undesired if assigning $i$ and $j$ to different labels yields low energy.
However, when variables $i$ and $j$ are {\em in agreement } under the energy (i.e., assignments with $l_i=l_j$ yield low energy),
aggregating them together allows for efficient exploration of low energy assignments.
{\bf A desired interpolation aggregates $i$ and $j$ when $i$ and $j$ are in agreement under the energy}.


\subsection{Measuring energy\-/aware agreements}
\label{sec:local-correlations}
We provide two measures for agreement, one is used for computing variable-coarsening ($P$),
while the other is used for label coarsening ($\hat{P}$).

\noindent{\bf Energy-aware agreement between variables:}
A reliable estimation for the agreement between the variables allows us to construct a desirable $P$ that aggregates variables that are in agreement under the energy.
A na\"{\i}ve approach would assume that neighboring variables are always in agreement (this assumption underlies the diadic pyramids of \cite{Felzenszwalb2006,Komodakis2010}).
This assumption clearly does not hold in general and may yield an undesired interpolation matrix $P$ leading to an inefficient multiscale scheme.
More recently \cite{Kim2011} suggested to use the energy itself in order to estimate variable agreements.
However, their ad-hoc methods are incapable of balancing the effect of the unary and pair-wise terms of the energy.

Indeed it is difficult to decide which term dominates and how to fuse these two terms together.
Therefore, we propose a novel empirical scheme for agreement estimation that
naturally accounts for and integrates the influence of both the unary and the pair-wise terms.
Moreover, our method
applies to all energies (\ref{eq:EngU}): submodular, non\-/submodular, metric $V$, arbitrary $V$, arbitrary $W$, energies defined over regular grids and arbitrary graphs.

Variables $i$ and $j$ are in agreement under the energy when $l_i=l_j$ yields relatively low energy value.
To estimate these agreements we empirically generate several samples with relatively low energy,
and measure the label agreement between neighboring variables $i$ and $j$ in these samples.
We use Iterated Conditional Modes (ICM) of \cite{Besag1986} to obtain locally low energy assignments:
Starting with a random assignment ICM chooses, at each iteration, for each variable, the label
yielding the largest decrease of the energy function, conditioned on the labels assigned to its neighbors.

This procedure may be viewed as a special case of sampling from a distribution:
The assumed underlying distribution is a Gibbs distribution, i.e., $p\left(U\right)\propto\exp\left(-\frac{1}{T}E\left(U\right)\right)$.
ICM may be interpreted as Gibbs sampling from the distribution at the limit $T\rightarrow 0$ (i.e., the "zero-temperature" limit).
Therefore, our samples may be viewed as zero-temperature Gibbs sampling with multiple restarts from the posterior (\cite{Koller2009}).


Performing $t=10$ ICM iterations with $K=10$ random restarts provides us with $K$ samples $\left\{L^k\right\}_{k=1}^K$.
Utilizing the label-disagreement weights encoded in the matrix $V$,
the disagreement between neighboring variable $i$ and $j$ is estimated as $d_{ij}=\frac{1}{K}\sum_k V_{l^k_i,l^k_j}$, where $l^k_i$ is the label of variable $i$ in the $k^{th}$ sample.
Their agreement is then given by $c_{ij}=\exp\left(-\frac{d_{ij}}{\sigma}\right)$,
with $\sigma \propto \max V$.


\noindent{\bf Energy-aware agreement between labels:}
Agreements between labels are easier to estimate, since this information is explicit in the matrix $V$ that encodes the label-disagreement between any two labels.
Setting
$\hat{c}_{\alpha,\beta}\propto \left(\hat{V}_{\alpha,\beta}\right)^{-1}$,
we get a ``closed-form" expression for the agreements between labels.


\subsection{From agreements to interpolation}
\label{sec:amg-p}

Using our measure for the variable agreements, $c_{ij}$, we follow the Algebraic Multigrid (AMG) method of \cite{Brandt1986} to first determine the set of coarse representatives and then construct an interpolation matrix $P$ that softly aggregates variables according to their agreement.

We begin by selecting a set of coarse representative variables $\VV^c\subset \VV^f$,
such that every variable in $\VV^f \backslash \VV^c$ is in agreement with $\VV^c$.
A variable $i$ is considered in agreement with $\VV^c$ if $\sum_{j\in\VV^c}c_{ij} \ge \beta \sum_{j\in\VV^f} c_{ij}$.
That is, every variable in $\VV^f$ is either in $\VV^c$ or is {\em in agreement} with other variables in $\VV^c$,
and thus well represented in the coarse scale.

We perform this selection greedily and sequentially, starting with $\VV^c=\emptyset$ adding $i$ to $\VV^c$ if it is not yet in agreement with $\VV^c$.
The parameter $\beta$ affects the coarsening rate, i.e., the ratio $n^c/n^f$,
smaller $\beta$ results in a lower ratio.

At the end of this process we have a set of coarse representatives $\VV^c$.
The interpolation matrix $P$ is then defined by:
\begin{equation}
P_{iI(j)} = \left\{
\begin{array}{cl}
c_{ij}                & i\in\VV^f\backslash\VV^c,\ j\in\VV^c\\
1                      & i\in\VV^c, j=i\\
0                      & \mbox{otherwise}\\
\end{array}
\right. \label{eq:entries-of-P}
\end{equation}
Where $I(j)$ is the coarse index of the variable whose fine index is $j$ (in Fig.~\ref{fig:multiscale}: $I(2)=1$ and $I(3)=2$).

We further prune rows of $P$ leaving only $\delta$ maximal entries.
Each row is then normalized to sum to 1.
Throughout our experiments we use $\beta=0.2$ ($\hat{\beta}=0.75$), $\delta=3$ ($\hat{\delta}=2$) for computing $P$ ($\hat{P}$ resp.).

\section{A Unified Discrete Multiscale Framework}
\label{sec:pipeline}

\begin{algorithm}[t]
\caption{Discrete multiscale optimization. \label{alg:multiscale}}
\DontPrintSemicolon
\SetKw{KwInit}{Init}
\SetKw{KwOpt}{Refine}
\SetKw{KwCoarse}{Coarsen}
\KwIn{Energy $\left(n^0, l, D^0, W^0, V\right)$.}
\KwOut{$U^0$}
\KwInit{$s\leftarrow 0$}\tcp{fine scale}
\tcp{Energy pyramid construction:}
\While{$\abs{\VV^s} \ge 10$} {
    Estimate pair-wise agreements $c_{ij}$ at scale $s$ (Sec.~\ref{sec:local-correlations}).\;
    Compute interpolation matrix $P^s$ (Sec.~\ref{sec:amg-p}).\;
    Derive coarse energy $\left(n^{s+1}, l, D^{s+1}, W^{s+1}, V\right)$ (Eq.~\ref{eq:EngC}).\;
    $s++$\;
}
\tcp{Coarse-to-fine optimization:}
\While{$s\ge0$} {
    $U^s\leftarrow$ \KwOpt{$(\tilde{U}^s)$}\;
    $\tilde{U}^{s-1} = P^sU^s$\tcp{interpolate a solution}\label{line:refine}
    $s--$\;
}
where \KwOpt{$(\tilde{U}^s)$} uses an existing single\-/scale method to optimize the energy $\left(n^{s}, l, D^{s}, W^{s}, V\right)$ with $\tilde{U}^s$ as an initialization.\;
\end{algorithm}

So far we have described the different components of our multiscale framework.
Alg.~\ref{alg:multiscale} puts them together into a multiscale minimization scheme.
Given an energy $\left(n, l, D, W, V\right)$,
our framework first works fine-to-coarse to compute interpolation matrices $\left\{P^s\right\}$ that construct the ``energy pyramid": $\left\{\left(n^s, l, D^s, W^s, V\right)\right\}_{s=0,\ldots,S}$.
Typically we end up at the coarsest scale with less than $10$ variables.
As a result, exploring the energy at this scale is robust to the initial assignment of the single\-/scale method used\footnote{In practice, at the coarsest scale we use ``winner-take-all" initialization as suggested by \cite[\S3.1]{Szeliski2008}.}.

Starting from the coarsest scale, we apply a simple single\-/scale optimization method (e.g., ICM, $\alpha$-expansion, etc.).
Since there are very few degrees of freedom at the coarsest scale, these single\-/scale methods are likely to obtain a low-energy coarse solution.
This stems from the fact that at the coarsest scale the large basins of attraction of the energy are easily accessed and explored.

At each scale $s$, the coarse solution $U^s$ is interpolated to a finer scale $s-1$: $\tilde{U}^{s-1} \leftarrow P^sU^s$.
At the finer scale $\tilde{U}^{s-1}$ serves as a good initialization for optimizing the energy with the same single\-/scale optimization method.
These two steps of interpolation followed by refinement are repeated for all scales from coarse to fine.

Single-scale optimization methods for discrete energies generally accept only discrete assignments (i.e., the binary constraints~(\ref{eq:const-U})) as an initialization.
However, the interpolated solution $\tilde{U}^{s-1}$, at each scale, might not satisfy the binary constraints~(\ref{eq:const-U}).
Therefore, we round each row of $\tilde{U}^{s-1}$ by setting the maximal element to $1$ and the rest to $0$.

The most computationally intensive module
of our framework is the empirical estimation of the variable agreements.
The complexity of the agreement estimation is $O\left(\abs{\EE}\cdot l\right)$, where $\abs{\EE}$ is the number of non-zero elements in $W$ and $l$ is the number of labels.
However, it is fairly straightforward to parallelize this module.

It is now easy to see how our framework generalizes \cite{Felzenszwalb2006}, \cite{Komodakis2010} and \cite{Kim2011}.
They are restricted to hard aggregation in $P$.
\cite{Felzenszwalb2006} and \cite{Komodakis2010} use a multiscale pyramid, however their variable aggregation is not energy\-/aware, and is restricted to diadic pyramids.
On the other hand, \cite{Kim2011} have limited energy\-/aware aggregation, applied to two level ``pyramid" only.

\section{Experimental Results}
\label{sec:results}

We evaluated our multiscale framework on a diversity of discrete optimization tasks\footnote{code available at \url{www.wisdom.weizmann.ac.il/~bagon/matlab.html}.}: ranging from challenging non\-/submodular synthetic and co-clustering energies, to low-level submodular vision energies such as denoising and stereo.
In all of these experiments we minimize a {\em given} publicly available benchmark energy,
{\em we do not attempt to improve on the energy formulation itself}.

For every instance of energy minimization problem in these benchmarks we construct an energy pyramid using our method.
We then use our energy pyramid to efficiently exploit the multiscale landscape of each energy to improve optimization results of existing methods.
In the following experiments we use ICM (\cite{Besag1986}), $\alpha\beta$-swap and $\alpha$-expansion (large move making algorithms of \cite{Veksler2002}) as representative single\-/scale primal optimization algorithms.
Each step of the large move making algorithms of \cite{Veksler2002} solves a reduced binary problem.
For the challenging non\-/submodular energies these binary steps are approximated using QPBO(I) of \cite{Rother2007}.

We follow the protocol of \cite{Szeliski2008} that uses the {\em lower bound} of TRW-S (\cite{Kolmogorov2006}) as a baseline for comparing performance of different optimization methods on different energies.
We report the ratio between the resulting energy value and the lower bound
(in percents),
{\bf closer to $100\%$ is better}.

These experiments show how our energy\-/aware construction of the pyramid efficiently exposes the underlying multiscale landscape of the energy.
This way even simple and very local optimization scheme (applied at each scale) can achieve good approximations.
The most prominent example is ICM (\cite{Besag1986}): this greedy local coordinate descend algorithm performs
poorly when applied directly to the energy.
It converges very rapidly to a sub-optimal local solution (see, e.g., \cite{Szeliski2008}).
However, when used within our multiscale framework, local search at coarse scales amounts to very large and non-local search in the fine scale.
This example stresses the advantage of constructing energy\-/aware multiscale framework:
Exposing the multiscale landscape of the energy helps to achieve good approximation even when using simple and local methods at each scale.

When incorporating large move making algorithms as the single\-/scale optimization in our framework,
there is a  consistent improvement of multiscale over these single\-/scale scheme.
In addition, TRW-S is a dual method and is considered state\-/of\-/the\-/art for discrete energy minimization (\cite{Szeliski2008}).
However, we show that when it comes to non\-/submodular energies it struggles behind the large move making algorithms and even ICM.
Moreover, for these challenging energies, our multiscale framework gives a significant boost in optimization performance, achieving significantly lower energy values than the TRW-S.

\begin{table}
\caption{ {\bf Synthetic results:}
{\em
Showing percent of achieved energy value relative to the lower bound (closer to $100\%$ is better) for ICM, $\alpha\beta$-swap, $\alpha$-expansion and TRW-S
for varying strengths of the pair-wise term ($\lambda=5,10,15$, stronger $\rightarrow$ harder to optimize.)}
}
\centering
\setlength{\tabcolsep}{1mm}
\begin{tabular}{c||c|c||c|c||c|c||c}
 \multirow{3}{*}{$\lambda$} & \multicolumn{2}{c||}{ICM} & \multicolumn{2}{c||}{Swap(QPBO)} & \multicolumn{2}{c||}{Expand(QPBO)} & \multirow{2}{*}{TRW-S} \\
 & \multirow{2}{*}{{\color{ours}Ours}}  & single  &  \multirow{2}{*}{{\color{ours}Ours}} &single & \multirow{2}{*}{{\color{ours}Ours}}  & single & \\
 & & scale & & scale & & scale & \\\hline \hline
$5$ & {\color{ours}$112.6\%$} & $115.9\%$ & {\color{ours}$108.9\%$} & $110.0\%$ & {\color{ours}$110.5\%$} & $110.0\%$ & $116.6\%$ \\
$10$ & {\color{ours}$123.6\%$} & $130.2\%$ & {\color{ours}$118.5\%$} & $120.2\%$ & {\color{ours}$121.5\%$} & $121.0\%$ & $134.6\%$ \\
$15$ & {\color{ours}$127.1\%$} & $135.8\%$ & {\color{ours}$122.1\%$} & $124.1\%$ & {\color{ours}$124.6\%$} & $125.1\%$ & $138.3\%$ \\
\end{tabular}
\setlength{\tabcolsep}{\origtabcolsep}
\label{tab:res-synthetic}
\end{table}

\subsection{Synthetic}
We begin with synthetic {\em non\-/submodular} energies defined over a 4-connected grid graph of size $50\times50$ ($n=2500$), and $l=5$ labels.
The unary term $D \sim \mathcal{N}\left(0,1\right)$.
The pair-wise term $V_{\alpha\beta}=V_{\beta\alpha} \sim \mathcal{U}\left(0, 1\right)$ ($V_{\alpha\alpha}=0$) and $w_{ij}=w_{ji} \sim \lambda \cdot \mathcal{U}\left(-1,1\right)$.
The parameter $\lambda$ controls the relative strength of the pair-wise term,
stronger (i.e., larger $\lambda$) results with energies more difficult to optimize (see \cite{Kolmogorov2006}).
Table~\ref{tab:res-synthetic} shows results, averaged over 100 experiments.

Using our multiscale framework to perform coarse\-/to\-/fine optimization of the energy yields significantly lower energies for all single\-/scale methods used (ICM, $\alpha$-expansion and $\alpha\beta$-swap) and TRW-S:
The percents in {\color{ours}``ours"} column are closer to $100\%$ than the results of the other methods.

Despite the fact that these synthetic energies were randomly generated without any underlying structure,
still there is a multiscale landscape to the functional.
Our multiscale framework constructs an energy pyramid that exposes this underlying multiscale landscape,
resulting with better and more efficient optimization results.

The resulting synthetic energies are non\-/submodular (since $w_{ij}$ may become negative).
For these challenging energies, state-of-the-art dual method (TRW-S) performs rather poorly\footnote{We did not restrict the number of iterations, and let TRW-S run until no further improvement to the lower bound is made.} (worse than single scale ICM) and there is a significant gap between the lower bound and the energy of the actual primal solution provided.
This gap might be due to the fact that for these challenging no-submodular energies the dual bound is not tight (\cite{Werner2010}).

\begin{figure}
\centering
\newlength{\cipwidth}
\setlength{\cipwidth}{.115\linewidth}
\setlength{\tabcolsep}{0.5mm}
\begin{tabular}{c|c||c|c|c|c||c|c}
\multirow{3}{*}{GT} & \multirow{3}{*}{Input} & \multicolumn{2}{c|}{ICM} & \multicolumn{2}{c||}{QPBO} &\multirow{3}{*}{TRW-S} & Sim. \\
& & \multirow{2}{*}{{\color{ours} Ours}} & single & \multirow{2}{*}{{\color{ours} Ours}} & single &  & Ann.\\
& &                                      & scale  &                                      & scale  & & \\
\includegraphics[width=\cipwidth]{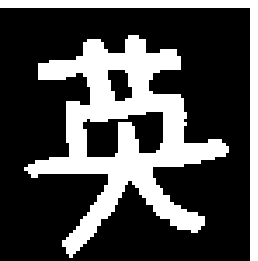}&
\includegraphics[width=\cipwidth]{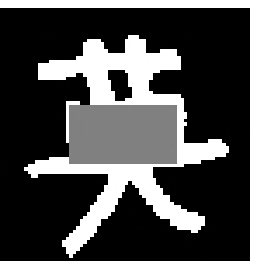}&
\includegraphics[width=\cipwidth]{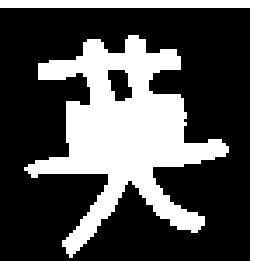}&
\includegraphics[width=\cipwidth]{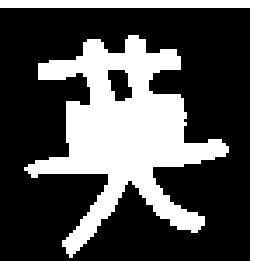}&
\includegraphics[width=\cipwidth]{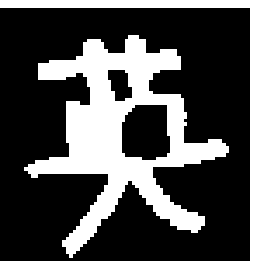}&
\includegraphics[width=\cipwidth]{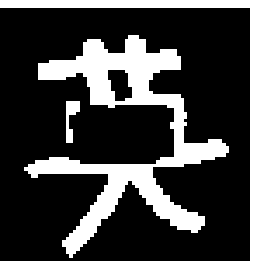}&
\includegraphics[width=\cipwidth]{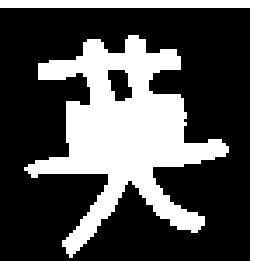}&
\includegraphics[width=\cipwidth]{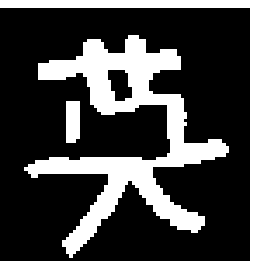}\\
\includegraphics[width=\cipwidth]{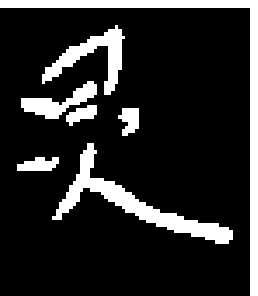}&
\includegraphics[width=\cipwidth]{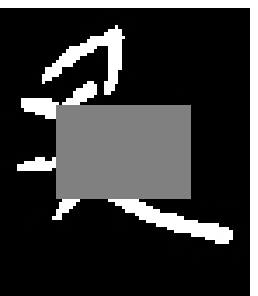}&
\includegraphics[width=\cipwidth]{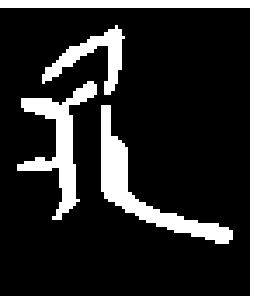}&
\includegraphics[width=\cipwidth]{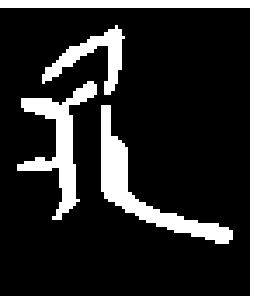}&
\includegraphics[width=\cipwidth]{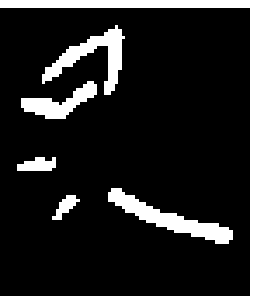}&
\includegraphics[width=\cipwidth]{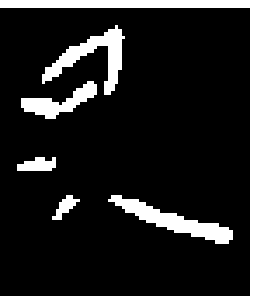}&
\includegraphics[width=\cipwidth]{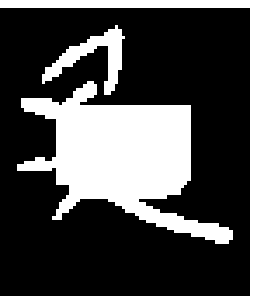}&
\includegraphics[width=\cipwidth]{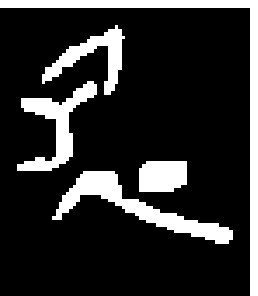}\\
\includegraphics[width=\cipwidth]{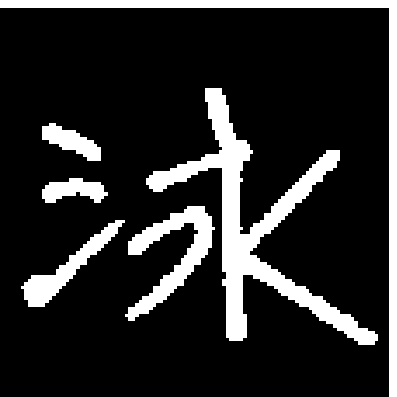}&
\includegraphics[width=\cipwidth]{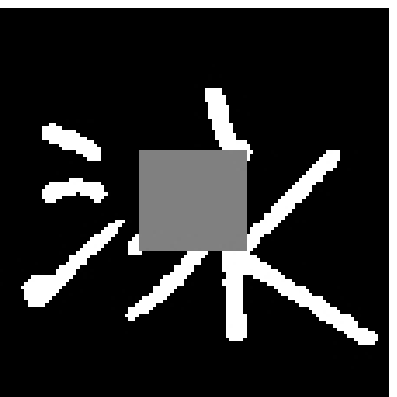}&
\includegraphics[width=\cipwidth]{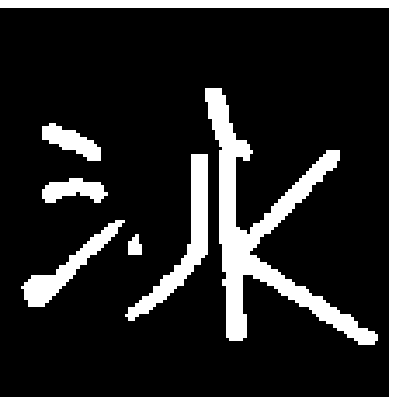}&
\includegraphics[width=\cipwidth]{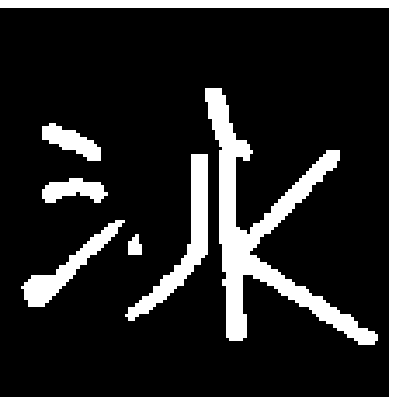}&
\includegraphics[width=\cipwidth]{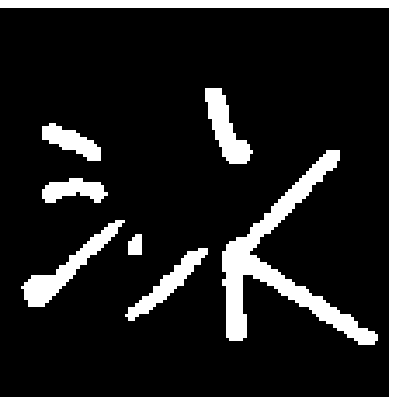}&
\includegraphics[width=\cipwidth]{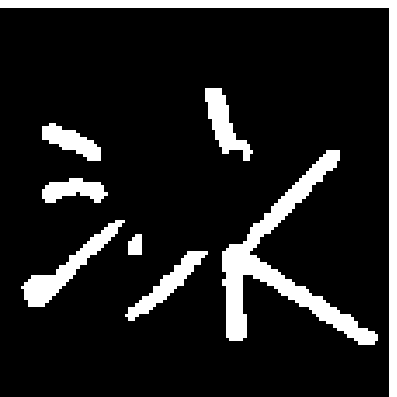}&
\includegraphics[width=\cipwidth]{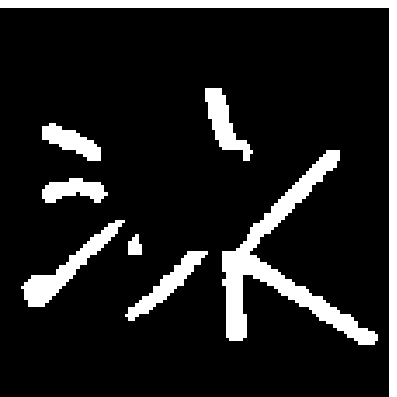}&
\includegraphics[width=\cipwidth]{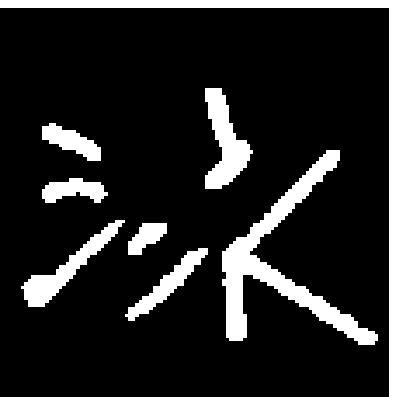}\\
\includegraphics[width=\cipwidth]{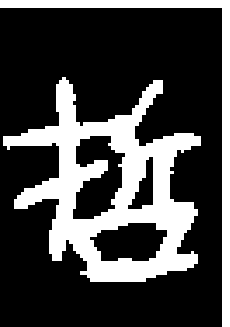}&
\includegraphics[width=\cipwidth]{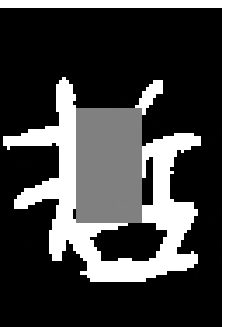}&
\includegraphics[width=\cipwidth]{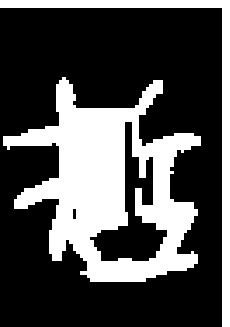}&
\includegraphics[width=\cipwidth]{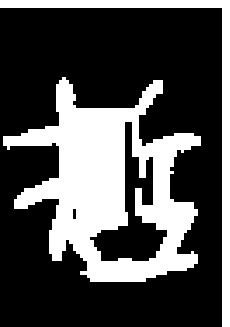}&
\includegraphics[width=\cipwidth]{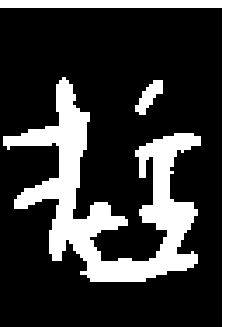}&
\includegraphics[width=\cipwidth]{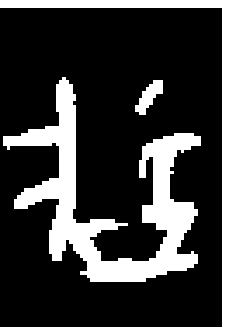}&
\includegraphics[width=\cipwidth]{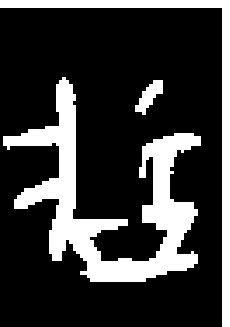}&
\includegraphics[width=\cipwidth]{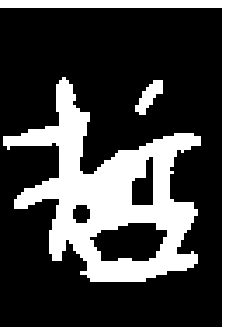}\\
\includegraphics[width=\cipwidth]{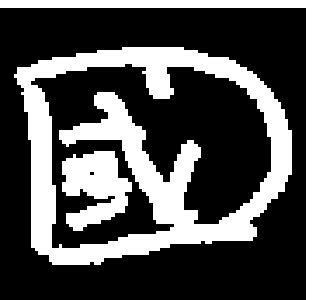}&
\includegraphics[width=\cipwidth]{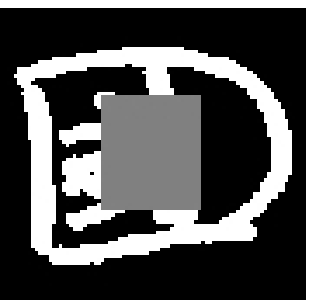}&
\includegraphics[width=\cipwidth]{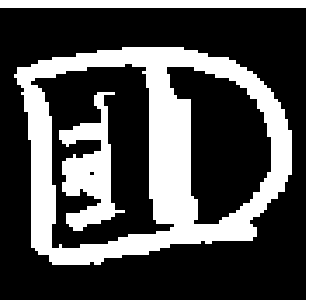}&
\includegraphics[width=\cipwidth]{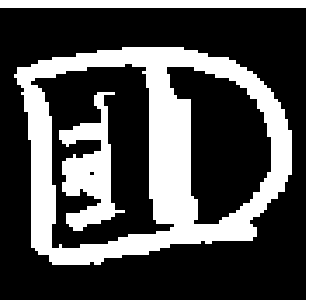}&
\includegraphics[width=\cipwidth]{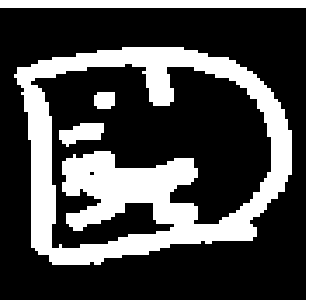}&
\includegraphics[width=\cipwidth]{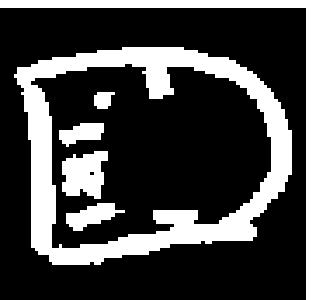}&
\includegraphics[width=\cipwidth]{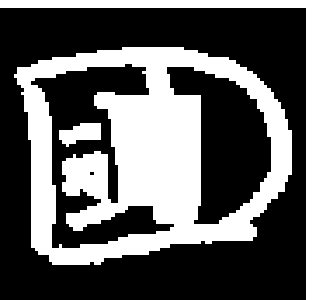}&
\includegraphics[width=\cipwidth]{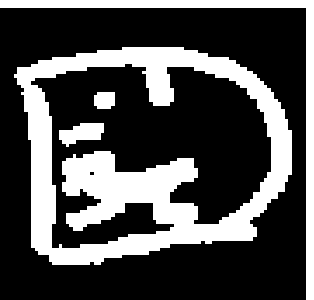}\\
\includegraphics[width=\cipwidth]{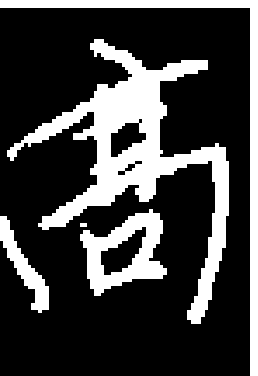}&
\includegraphics[width=\cipwidth]{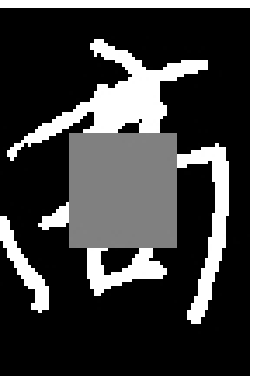}&
\includegraphics[width=\cipwidth]{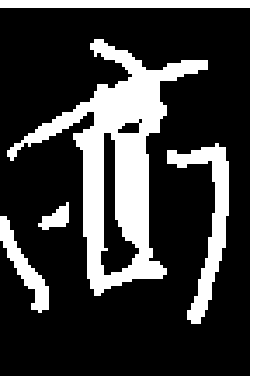}&
\includegraphics[width=\cipwidth]{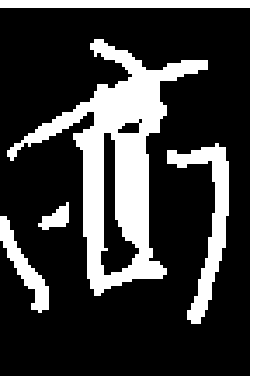}&
\includegraphics[width=\cipwidth]{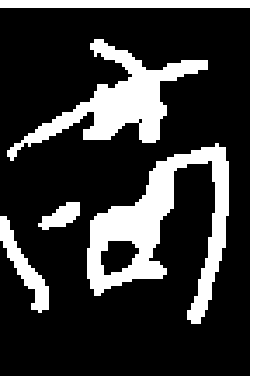}&
\includegraphics[width=\cipwidth]{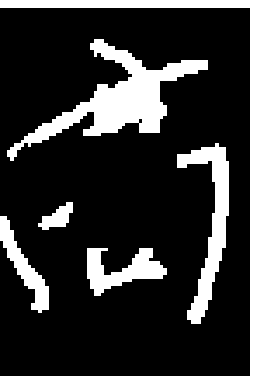}&
\includegraphics[width=\cipwidth]{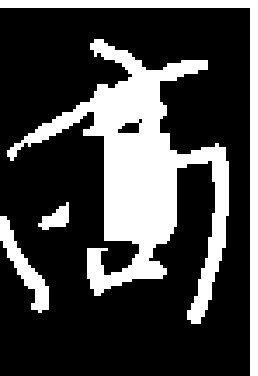}&
\includegraphics[width=\cipwidth]{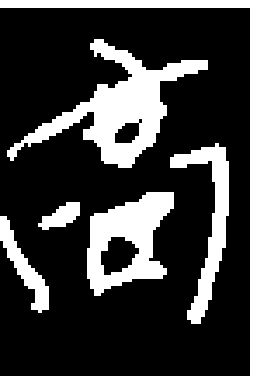}
\end{tabular}
\setlength{\tabcolsep}{\origtabcolsep}
\caption{{\bf Chinese characters inpainting:}
{\em Visualizing some of the instances used in our experiments.
Columns are (left to right):
The original character used for testing.
The input, partially occluded character.
ICM and QPBO results both our multiscale and single scale results.
Results of TRW-S and results of \protect\cite{Nowozin2011} obtained with a  very long run of simulated annealing (using Gibbs sampling inside the annealing).}}
\label{fig:sebastian-dtf}
\end{figure}

\begin{figure}
\centering
\includegraphics[width=\linewidth]{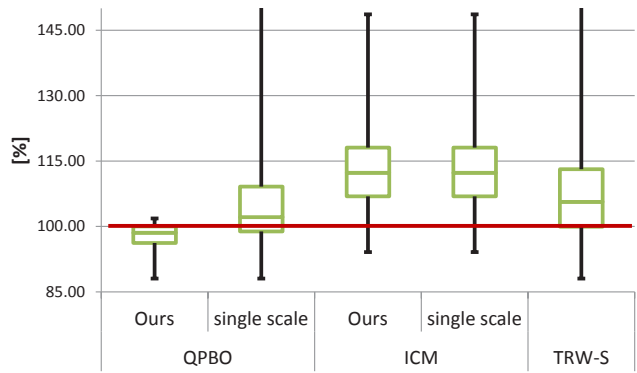}
\caption{ {\bf Energies of Chinese characters inpainting:}
{\em Box plot showing 25\%, median and 75\% of the resulting energies relative to reference energies of \protect\cite{Nowozin2011} (lower than $100\%$ = lower than baseline).
Our multiscale approach combined with QPBO achieves consistently better energies than baseline, with very low variance.
TRW-S improves on only 25\% of the instances with very high variance in the results.}
}
\label{fig:sebastian-dtf-graphs}
\end{figure}

\begin{table}
\centering
\caption{{\bf Energies of Chinese characters inpainting:}
{\em table showing
(a) mean energies for the inpainting experiment relative to baseline of \protect\cite{Nowozin2011} (lower is better, less than $100\%$ = lower than baseline).
(b) percent of instances for which strictly lower energy was achieved.
}}\label{tab:dtf-binary}
\begin{tabular}{c||c|c||c|c||c}
 & \multicolumn{2}{c||}{ICM} & \multicolumn{2}{c||}{QPBO} & \multirow{2}{*}{TRW-S} \\
 & {\color{ours}Ours} & single-scale & {\color{ours}Ours} & single-scale &  \\\hline
(a) &  {\color{ours}$114.0\%$} & $114.0\%$ & {\color{ours}$97.8\%$} & $106.2\%$ & $108.6\%$ \\\hline
(b) & {\color{ours}$7.0\%$} & $7.0\%$ &{\color{ours}$77.0\%$} & $34.0\%$ & $25.0\%$ \\\hline
\end{tabular}
\end{table}

\subsection{Chinese character inpainting}

We further experiment with non\-/submodular learned binary energies of \cite[\S5.2]{Nowozin2011}\footnote{available at \url{www.nowozin.net/sebastian/papers/DTF_CIP_instances.zip}.}.
These 100 instances of non\-/submodular pair-wise energies are defined over a 64-connected grid.
These energies were designed and trained to perform the task of learning Chinese calligraphy, represented as a complex, non-local binary pattern.

Our experiments show how approaching these challenging energies using our unified multiscale framework allows for better approximations.
Table~\ref{tab:dtf-binary} and Fig.~\ref{fig:sebastian-dtf} compare our multiscale framework to single\-/scale methods acting on the primal binary variables.
Since the energies are binary, multi-label large move making algorithms boils down to binary QPBO.
We also provide an evaluation of a dual method (TRW-S) on these energies.
In addition to the quantitative results, Fig.~\ref{fig:sebastian-dtf-graphs} provides a visualization of some of the instances of the restored Chinese characters.

For these challenging non\-/submodular `real world" energies our multiscale framework provides significant improvement over single\-/scale scheme.

\begin{table}
\caption{ {\bf Co-clustering results: }
{\em
Baseline for comparison are state-of-the-art results of \protect\cite{Glasner2011}.
(a) We report our results as percent of the baseline: smaller is better, lower than $100\%$ even outperforms state-of-the-art.
(b) We also report the fraction of energies for which our multiscale framework outperform state-of-the-art.
} }
\centering
\setlength{\tabcolsep}{.5mm}
\begin{tabular}{c||c|c||c|c||c|c||c}
 & \multicolumn{2}{c||}{ICM} & \multicolumn{2}{c||}{Swap(QPBO)} & \multicolumn{2}{c||}{Expand(QPBO)} & TRW-S\\
               & \multirow{2}{*}{{\color{ours}Ours}}     & single  & \multirow{2}{*}{{\color{ours}Ours}} & single &  \multirow{2}{*}{{\color{ours}Ours}} & single & \\
               &                          & scale &          & scale &   & scale &  \\
\hline \hline
(a) & {\color{ours}$99.9\%$} & $177.7\%$ & {\color{ours}$99.8\%$} & $101.5\%$ & {\color{ours}$99.8\%$} & $101.6\%$ & $176.2\%$ \\\hline
(b) & {\color{ours}$55.6\%$} & $0.0\%$ & {\color{ours}$71.8\%$} & $15.5\%$ & {\color{ours}$70.8\%$} & $11.6\%$ & $0.5\%$ \\\hline
\end{tabular}
\label{tab:cocluster-res}
\end{table}

\subsection{Co-clustering}

The problem of co-clustering addresses the matching of superpixels within and across frames in a video sequence.
Following \cite[\S6.2]{Bagon2012}, we treat co-clustering as a discrete minimization of {\em non\-/submodular} Potts energy.
We obtained 77 co-clustering energies, courtesy of \cite{Glasner2011}, used in their experiments.
The number of variables in each energy ranges from 87 to 788.
Their sparsity (percent of non-zero entries in $W$) ranges from $6\%$ to $50\%$,
The resulting energies are non\-/submodular, have no underlying regular grid, and are very challenging to optimize \cite{Bagon2012}.

Table~\ref{tab:cocluster-res} compares our discrete multiscale framework combined with ICM, $\alpha\beta$-swap and $\alpha$-expansion.
For these energies we use a different baseline: the state-of-the-art results of \cite{Glasner2011} obtained by applying specially tailored convex relaxation method
(We do not use the lower bound of TRW-S here since it is far from being tight for these challenging energies).
Our multiscale framework improves state-of-the-art for this family of challenging energies and significantly outperform TRW-S.

Furthermore, the results demonstrated in the last three sub-sections
highlight the advantage that primal methods has over dual ones when it comes to challenging non\-/submodular energies.

\subsection{Submodular energies}
We further applied our multiscale framework to optimize less challenging submodular energies.
We use the diverse low-level vision MRF energies from the Middlebury benchmark \cite{Szeliski2008}\footnote{Available at \url{vision.middlebury.edu/MRF/}.}.

\begin{figure*}
\centering
\newlength{\stwidth}
\setlength{\stwidth}{.12\linewidth}
\begin{tabular}{c@{ }c||c@{ }c||c@{ }c||c}
\multicolumn{2}{c||}{ICM} & \multicolumn{2}{c||}{Swap} & \multicolumn{2}{c||}{Expand} & Ground \\
{\color{ours}Ours} & Single scale & {\color{ours}Ours} & Single scale & {\color{ours}Ours} & Single scale & truth \\ \hline
\includegraphics[width=\stwidth]{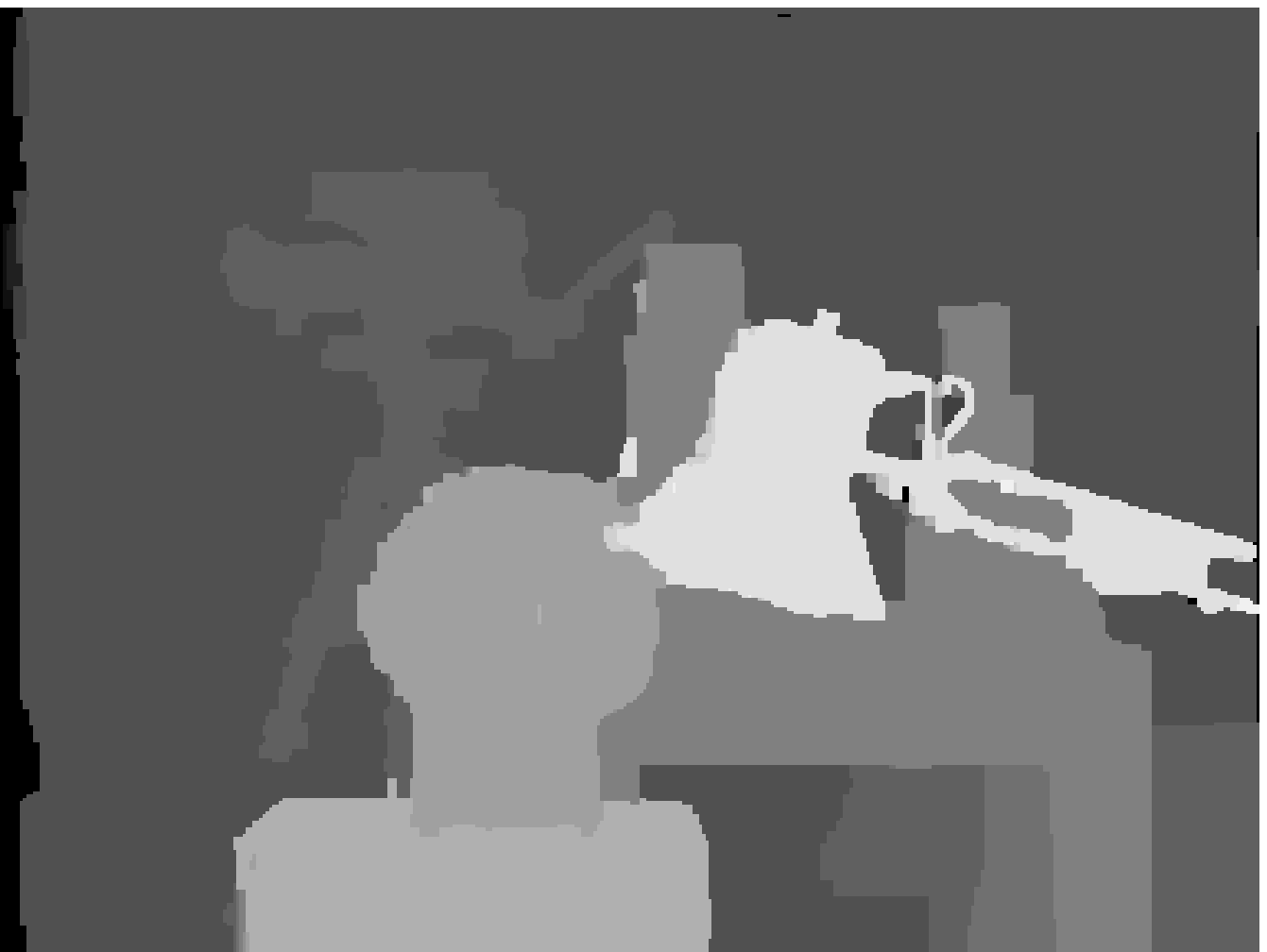}&
\includegraphics[width=\stwidth]{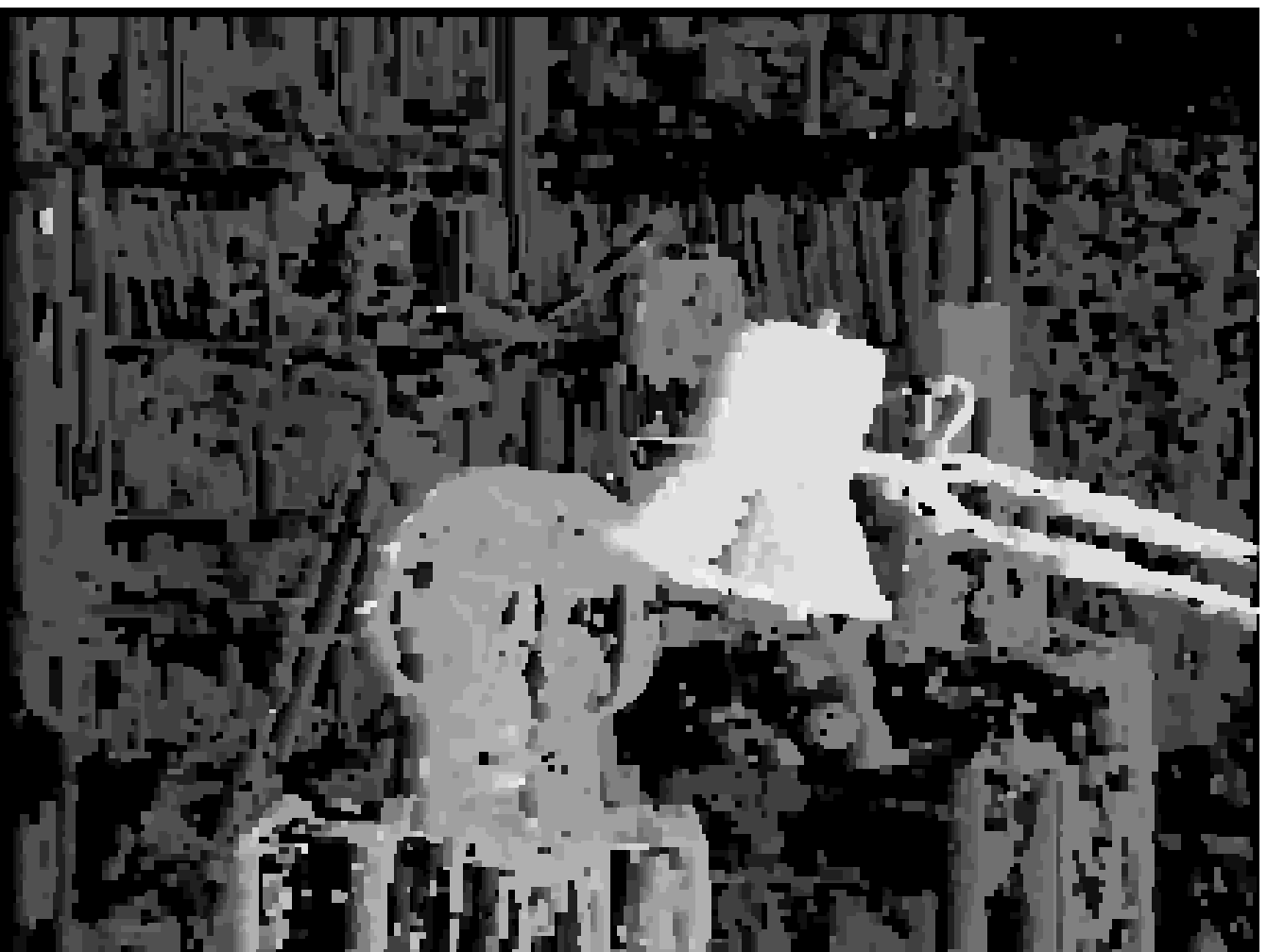}&
\includegraphics[width=\stwidth]{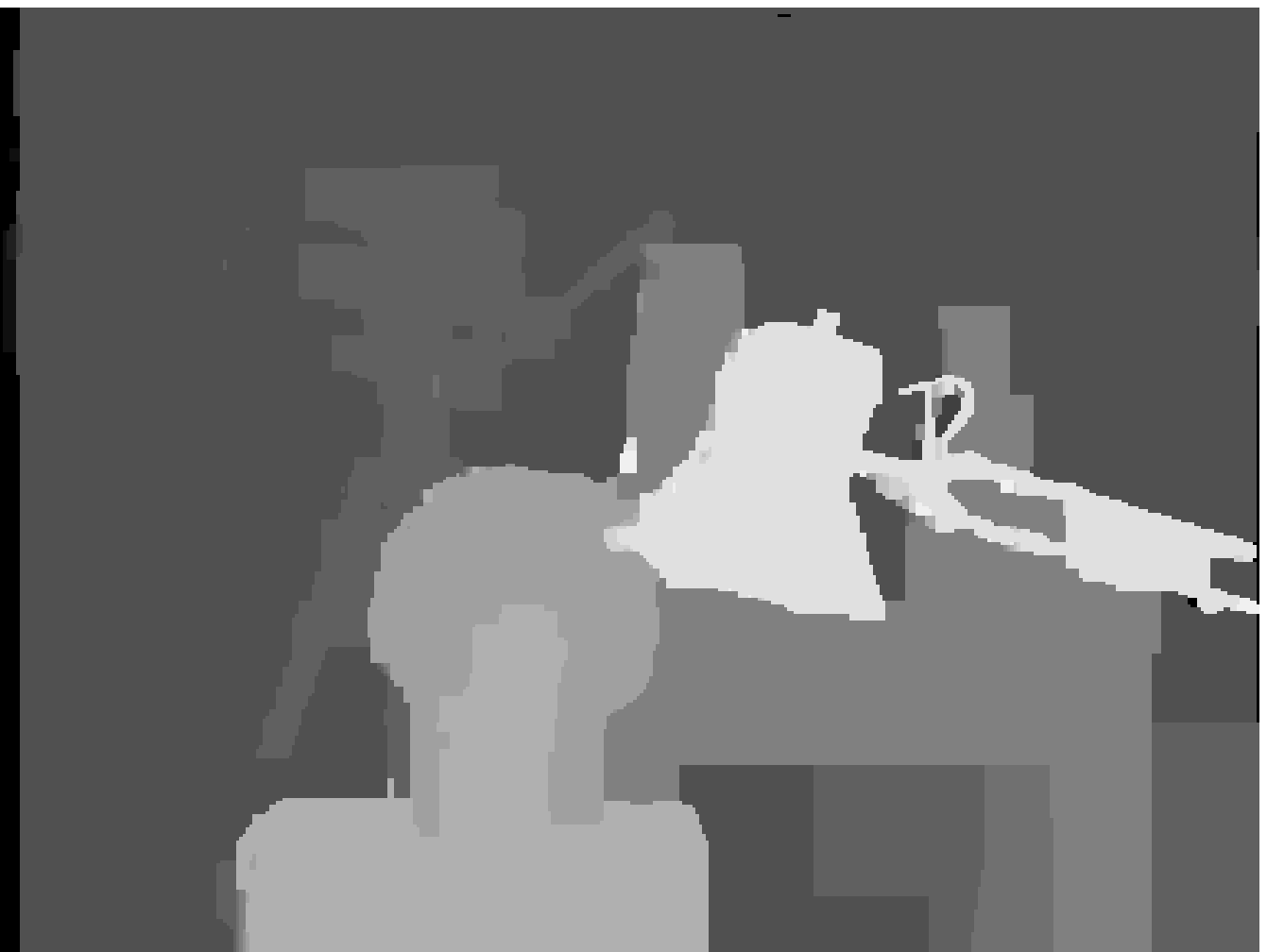}&
\includegraphics[width=\stwidth]{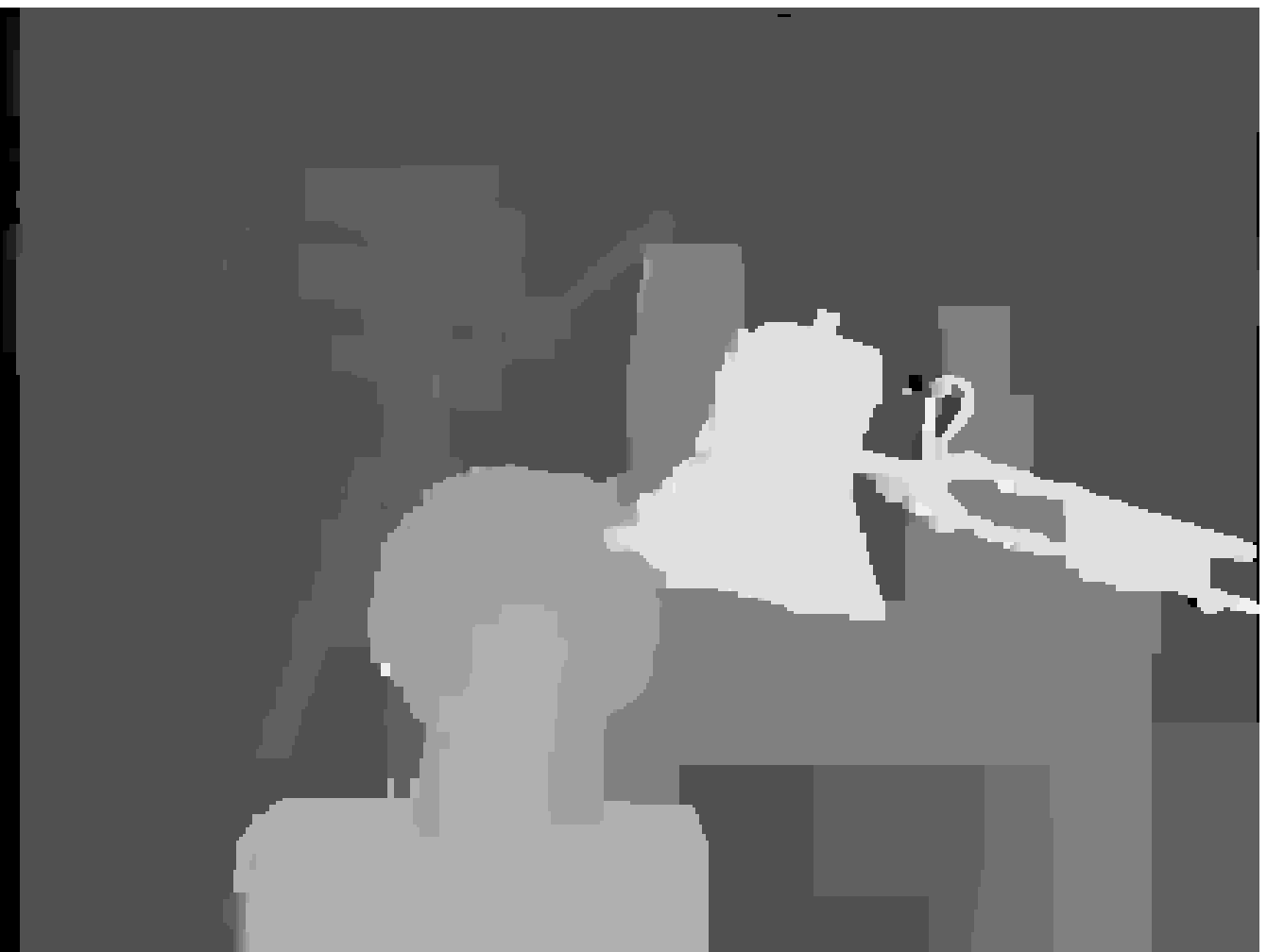}&
\includegraphics[width=\stwidth]{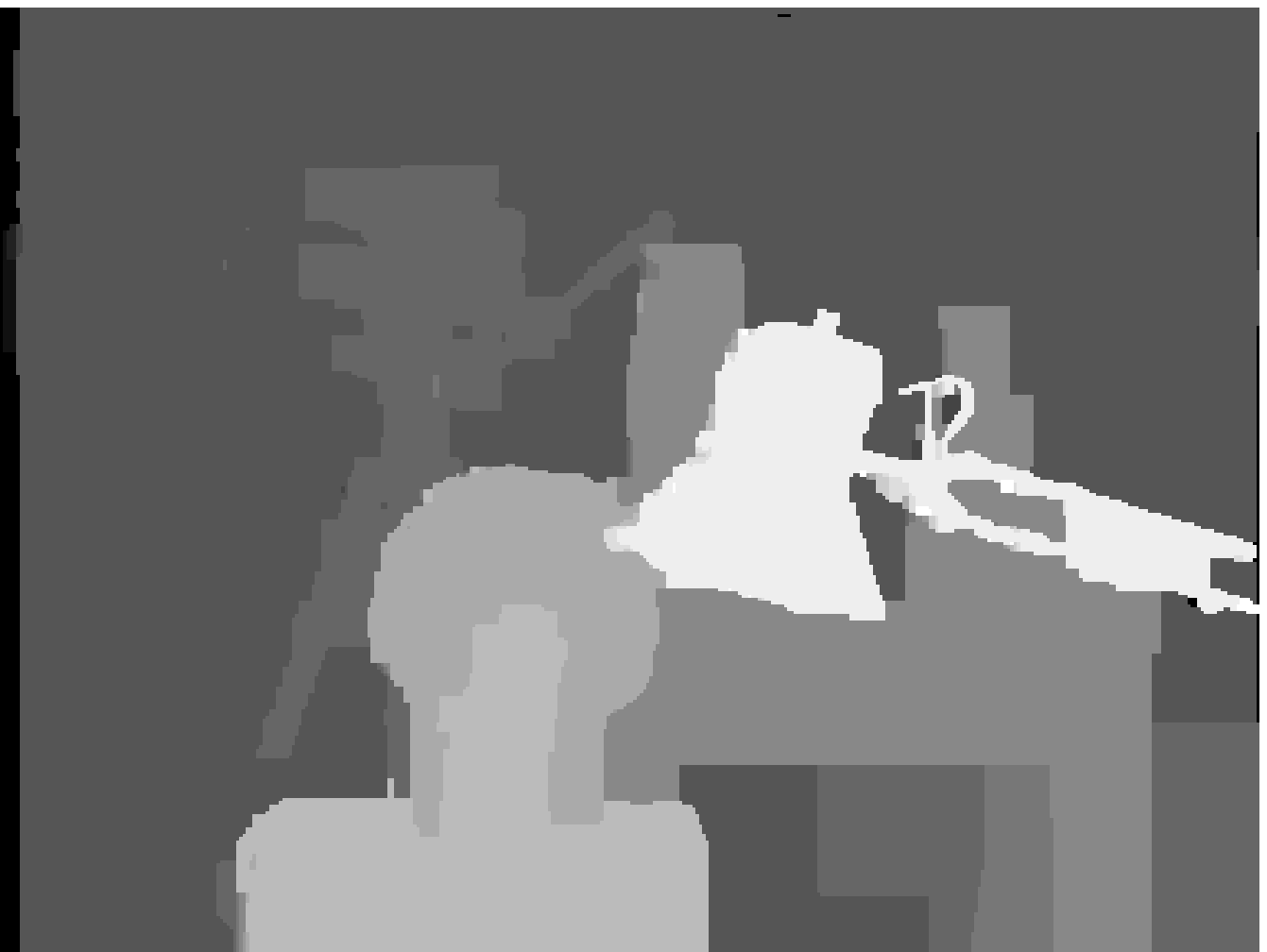}&
\includegraphics[width=\stwidth]{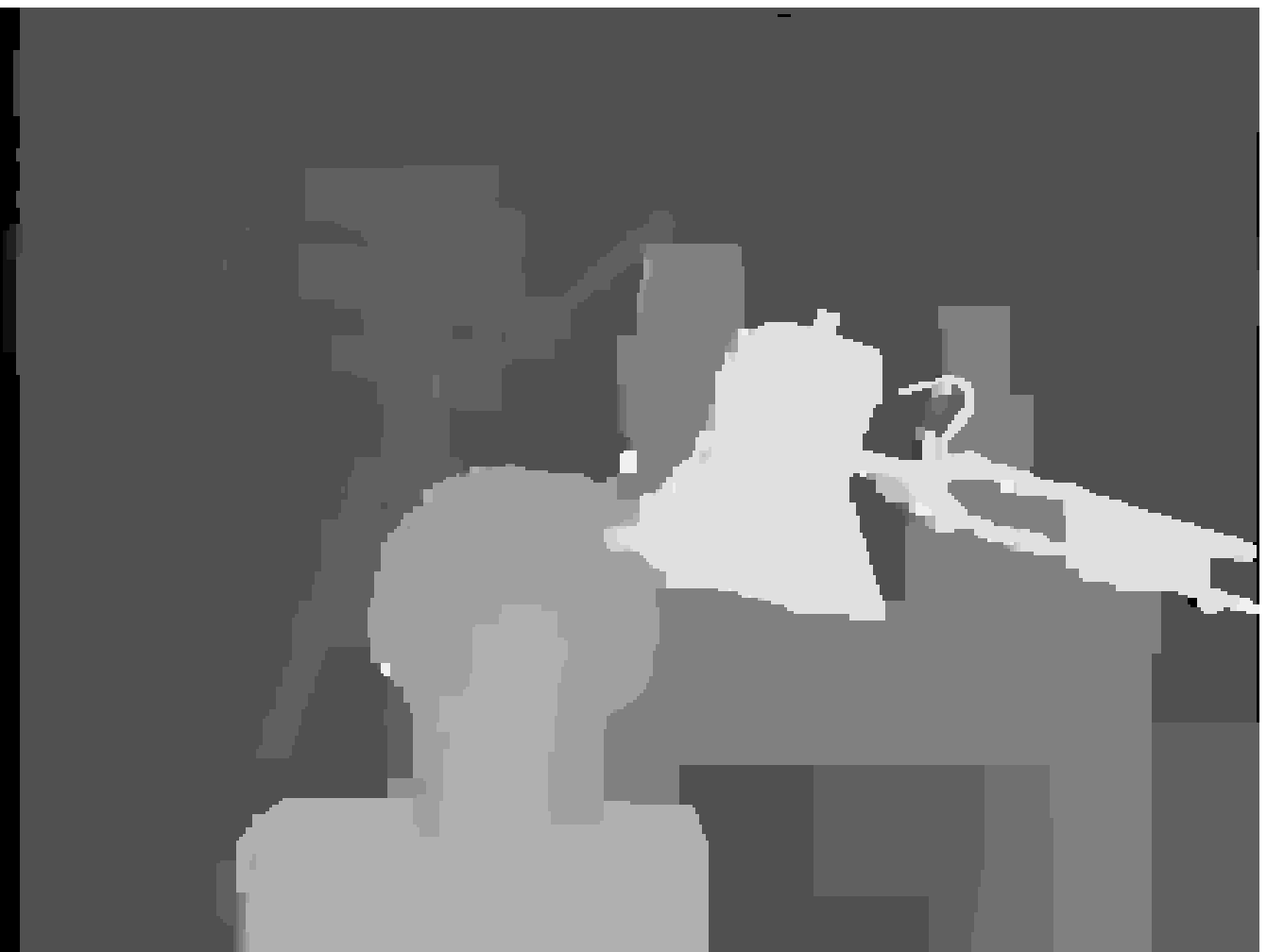}&
\includegraphics[width=\stwidth]{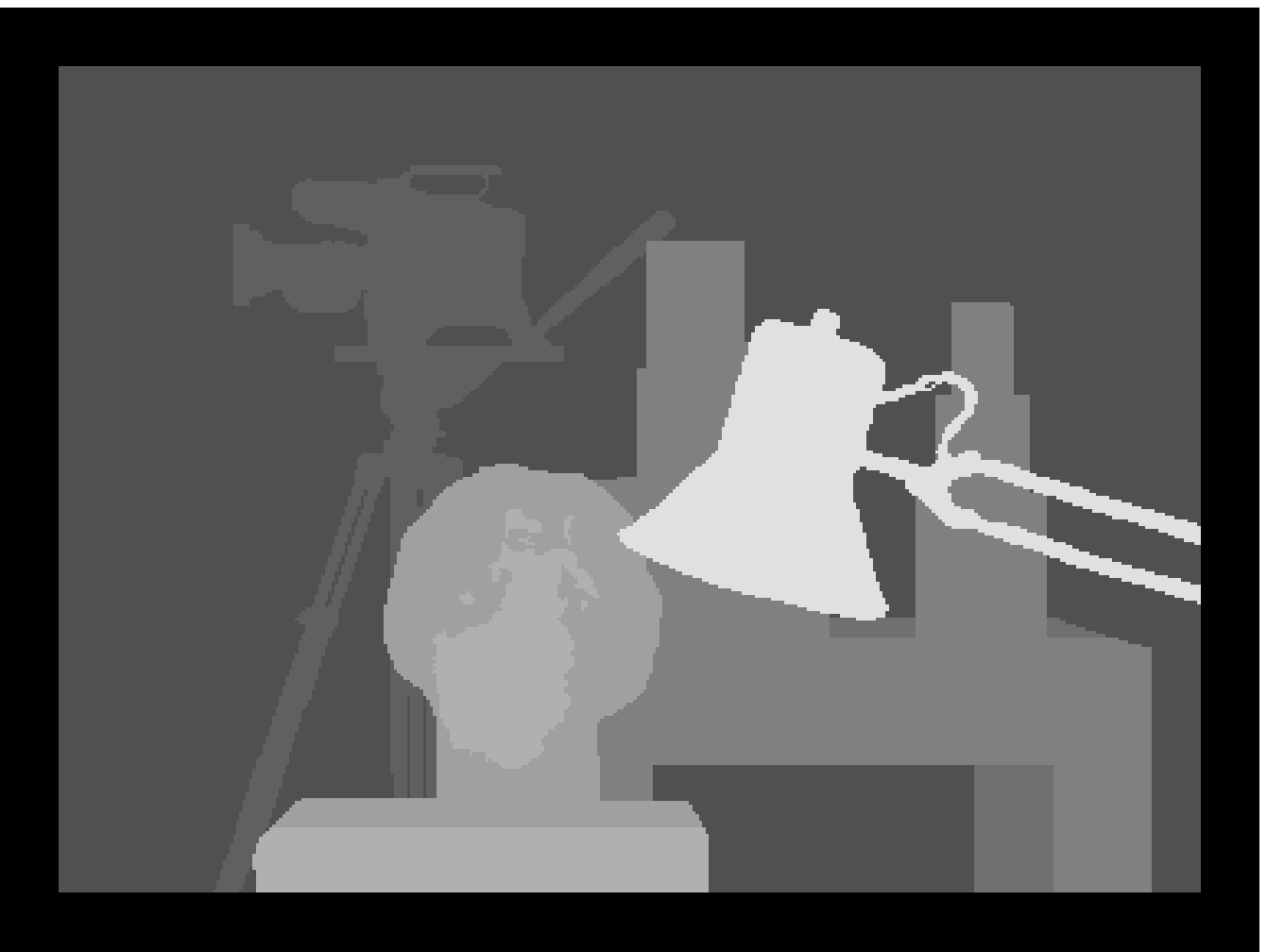}\\
\includegraphics[width=\stwidth]{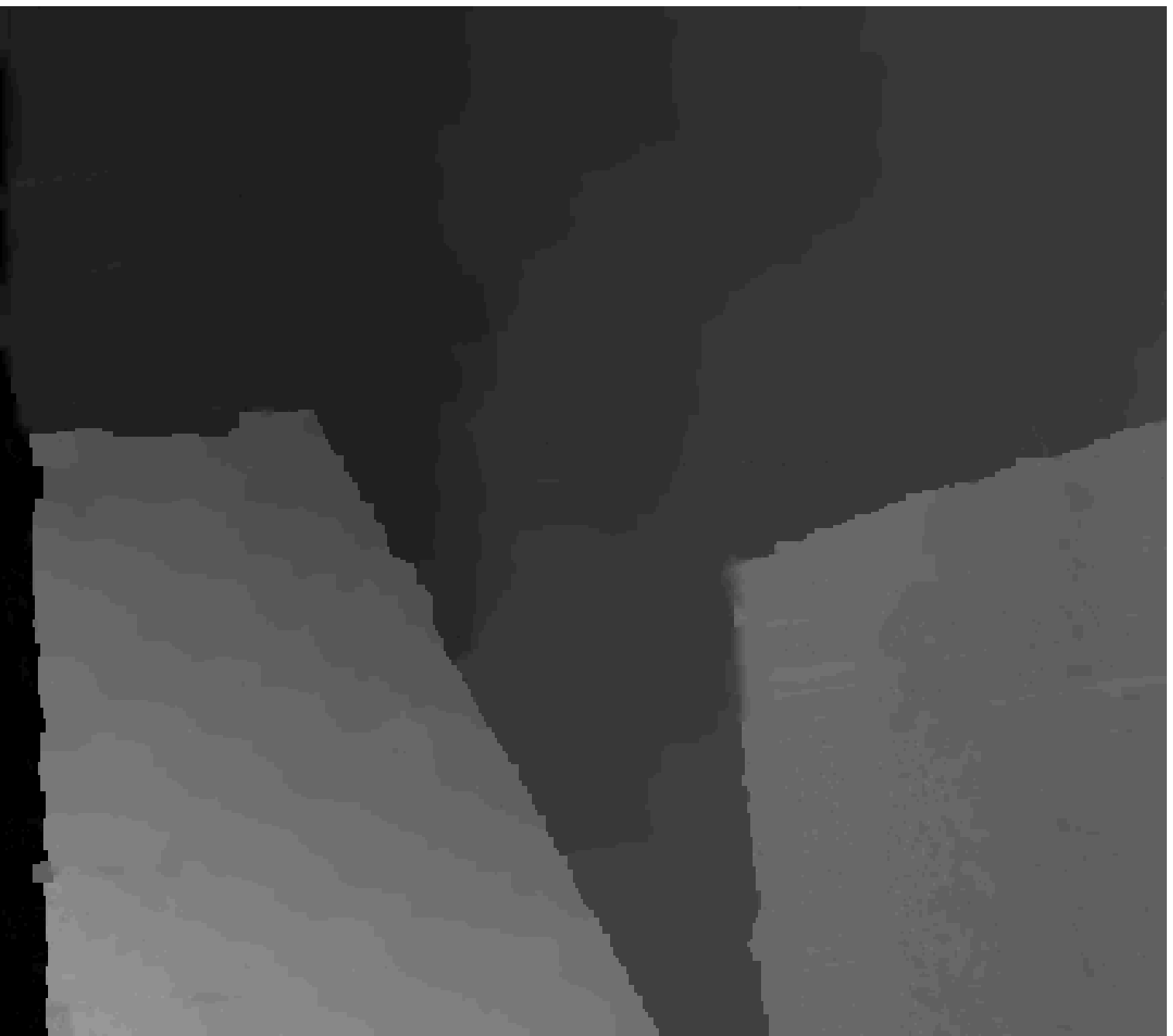}&
\includegraphics[width=\stwidth]{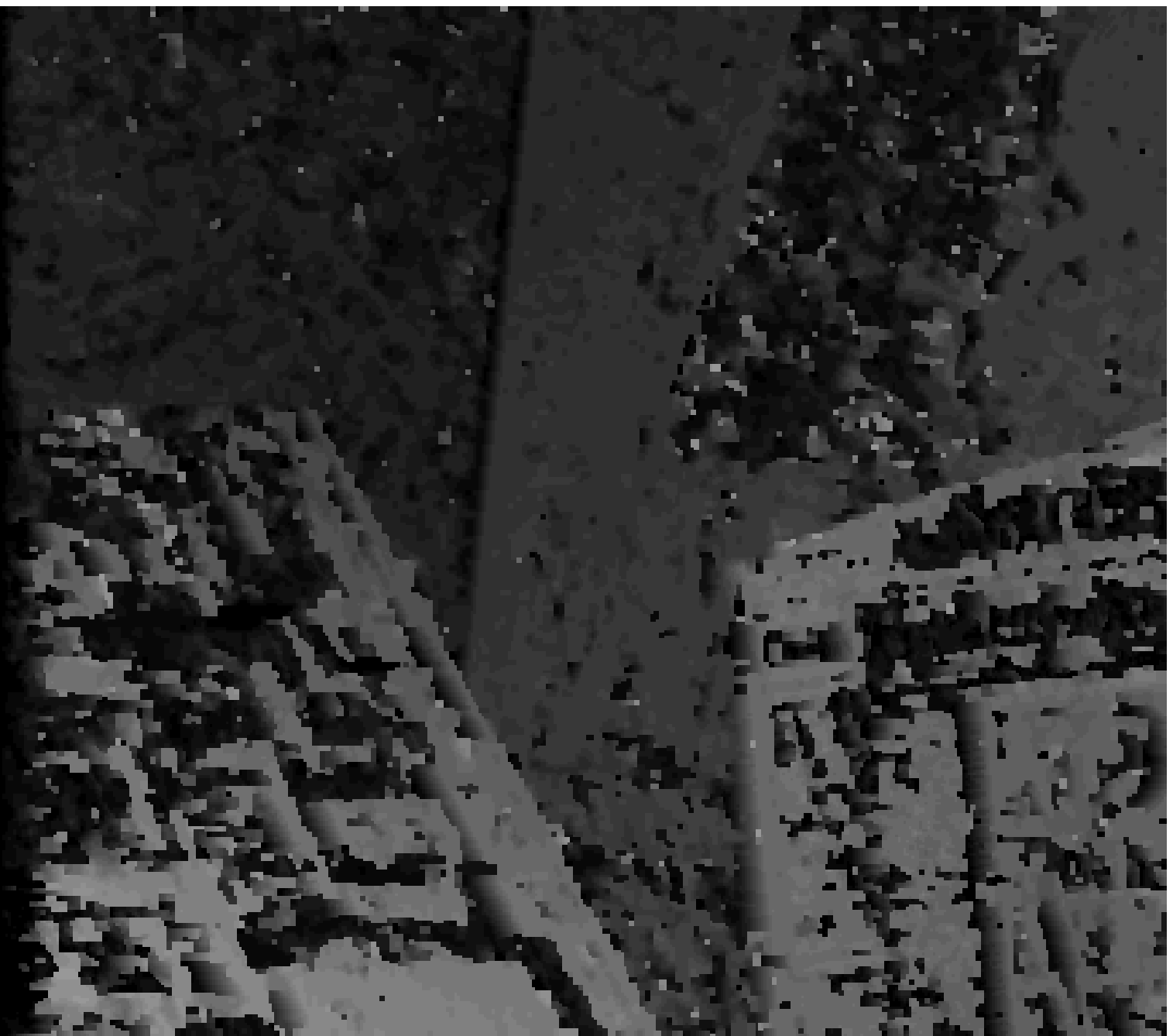}&
\includegraphics[width=\stwidth]{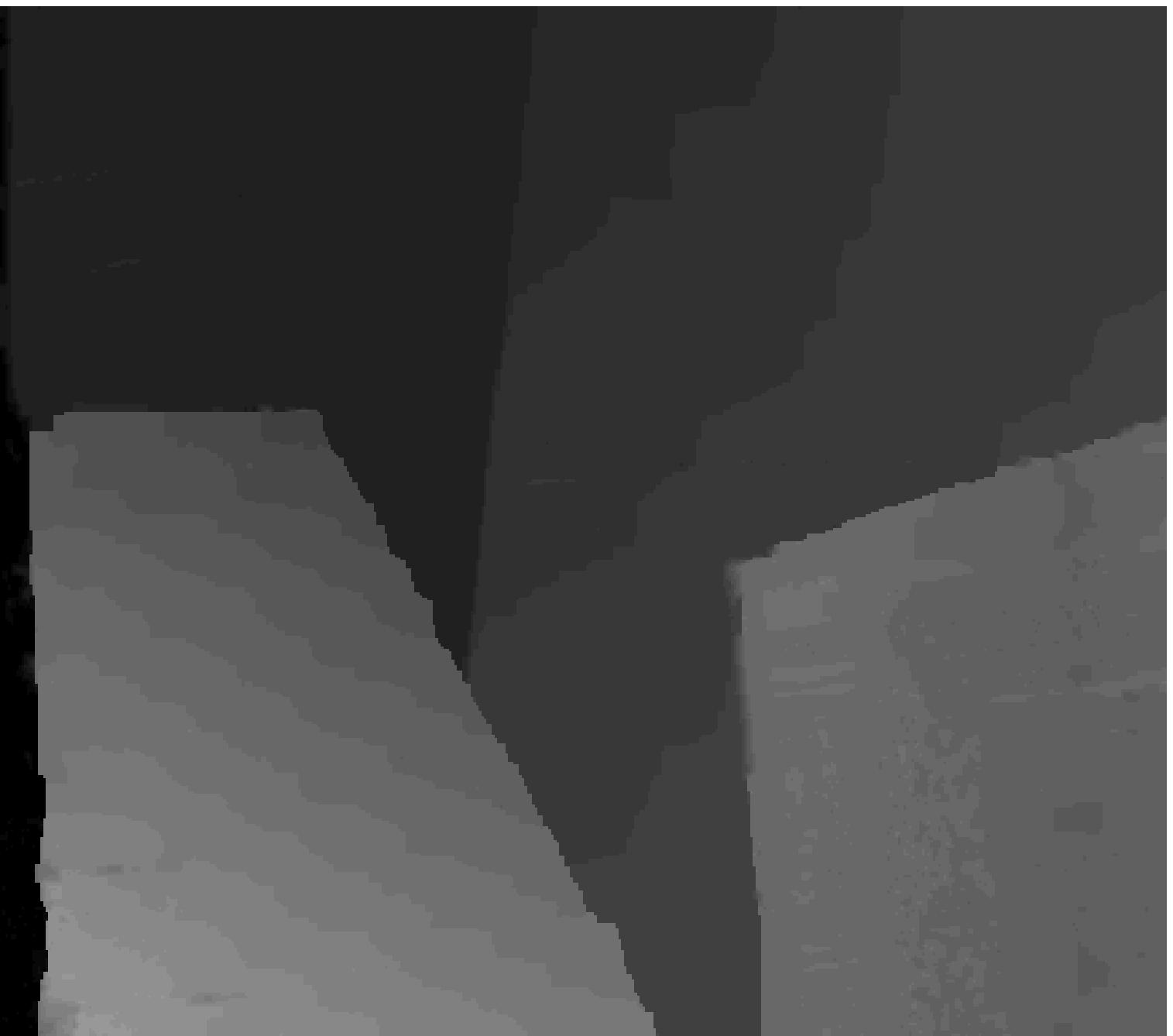}&
\includegraphics[width=\stwidth]{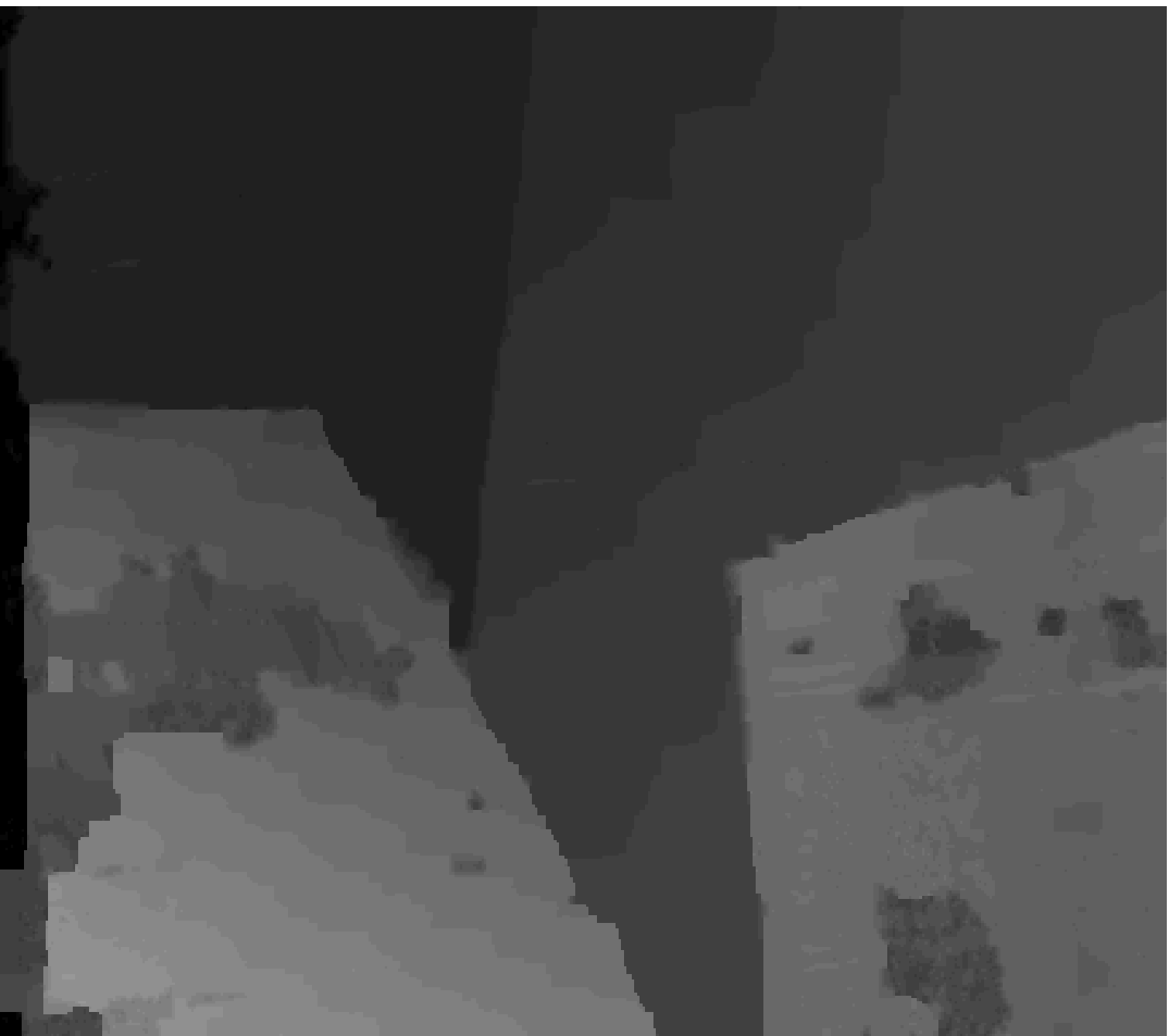}&
\includegraphics[width=\stwidth]{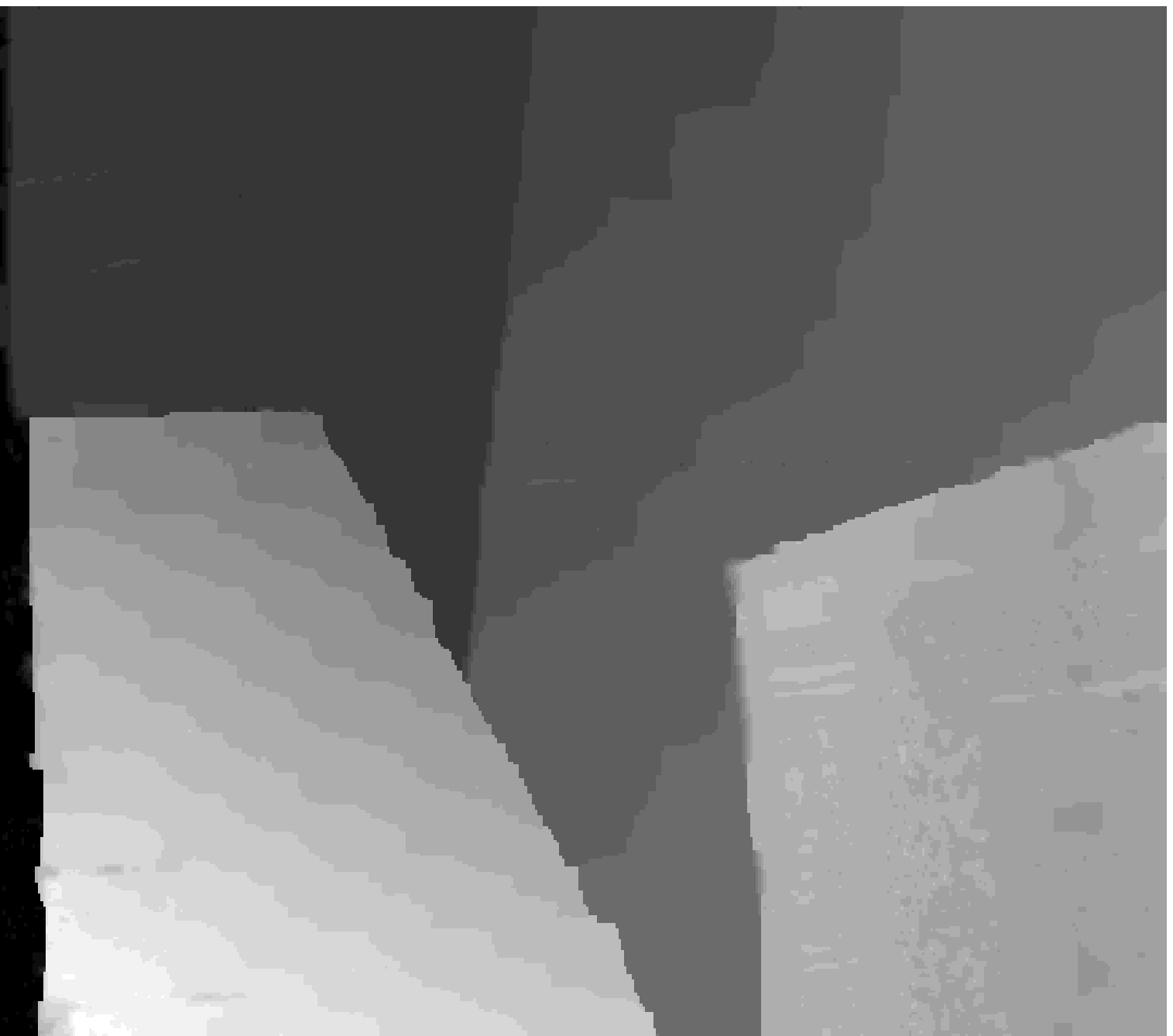}&
\includegraphics[width=\stwidth]{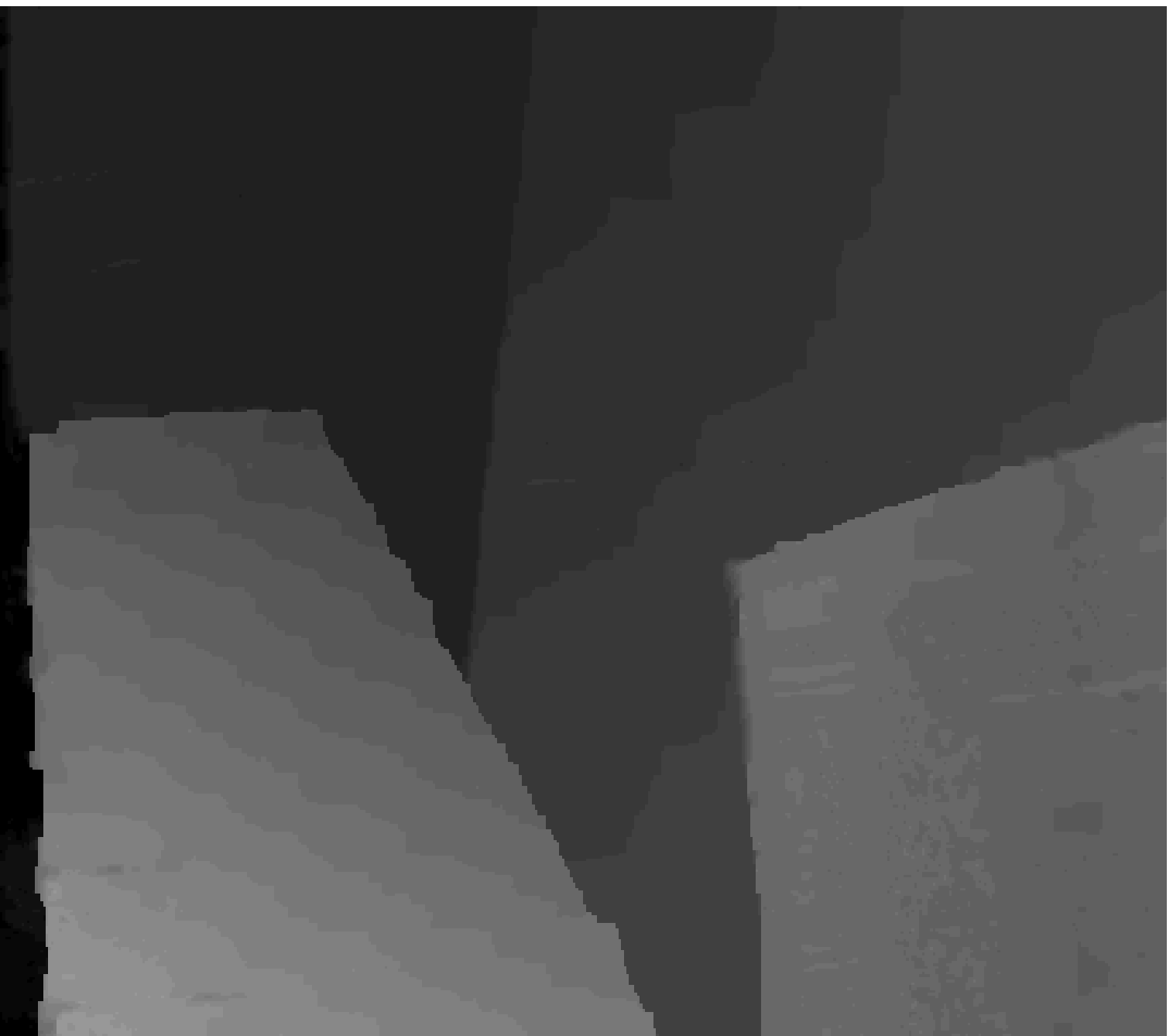}&
\includegraphics[width=\stwidth]{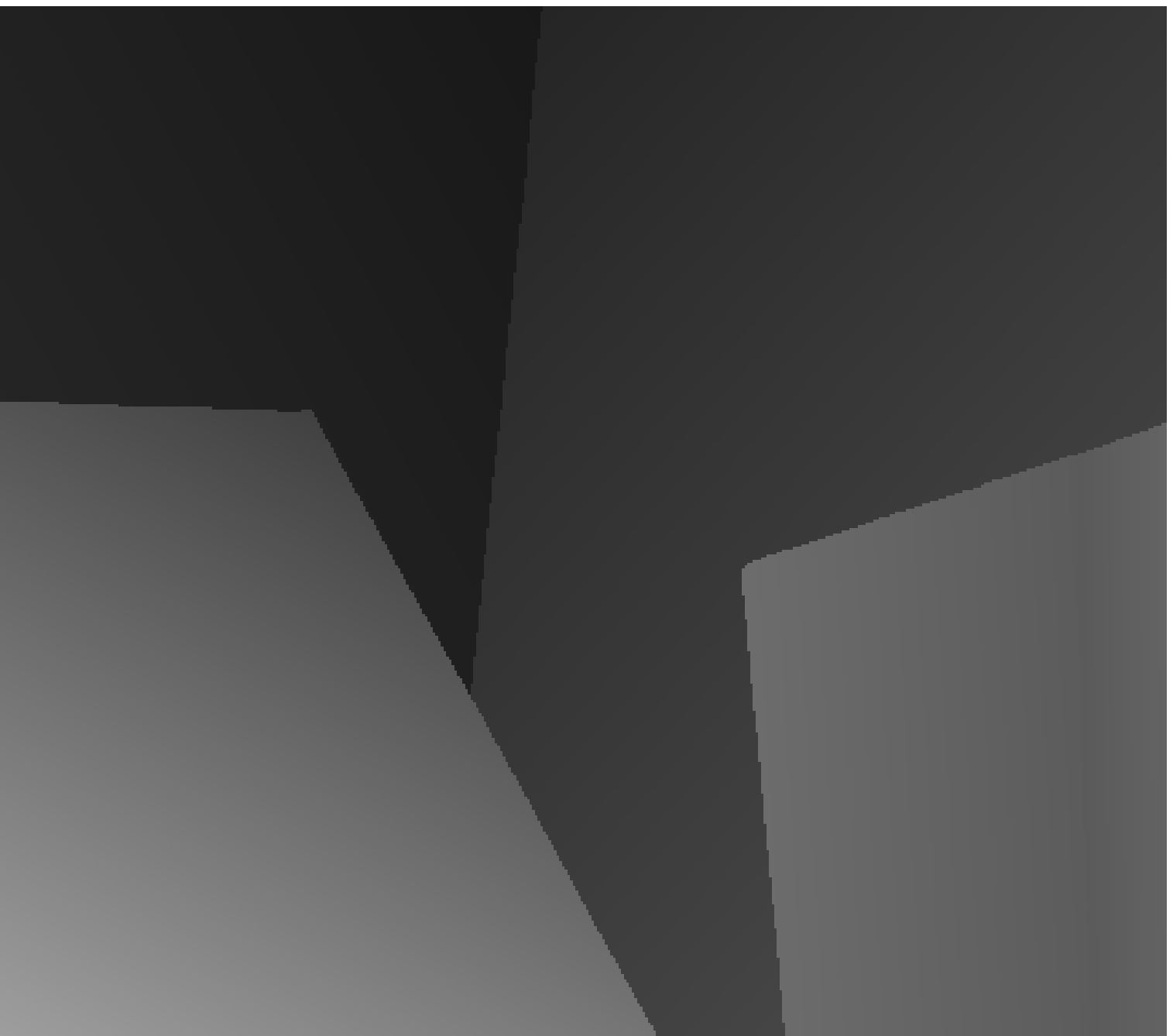}\\
\includegraphics[width=\stwidth]{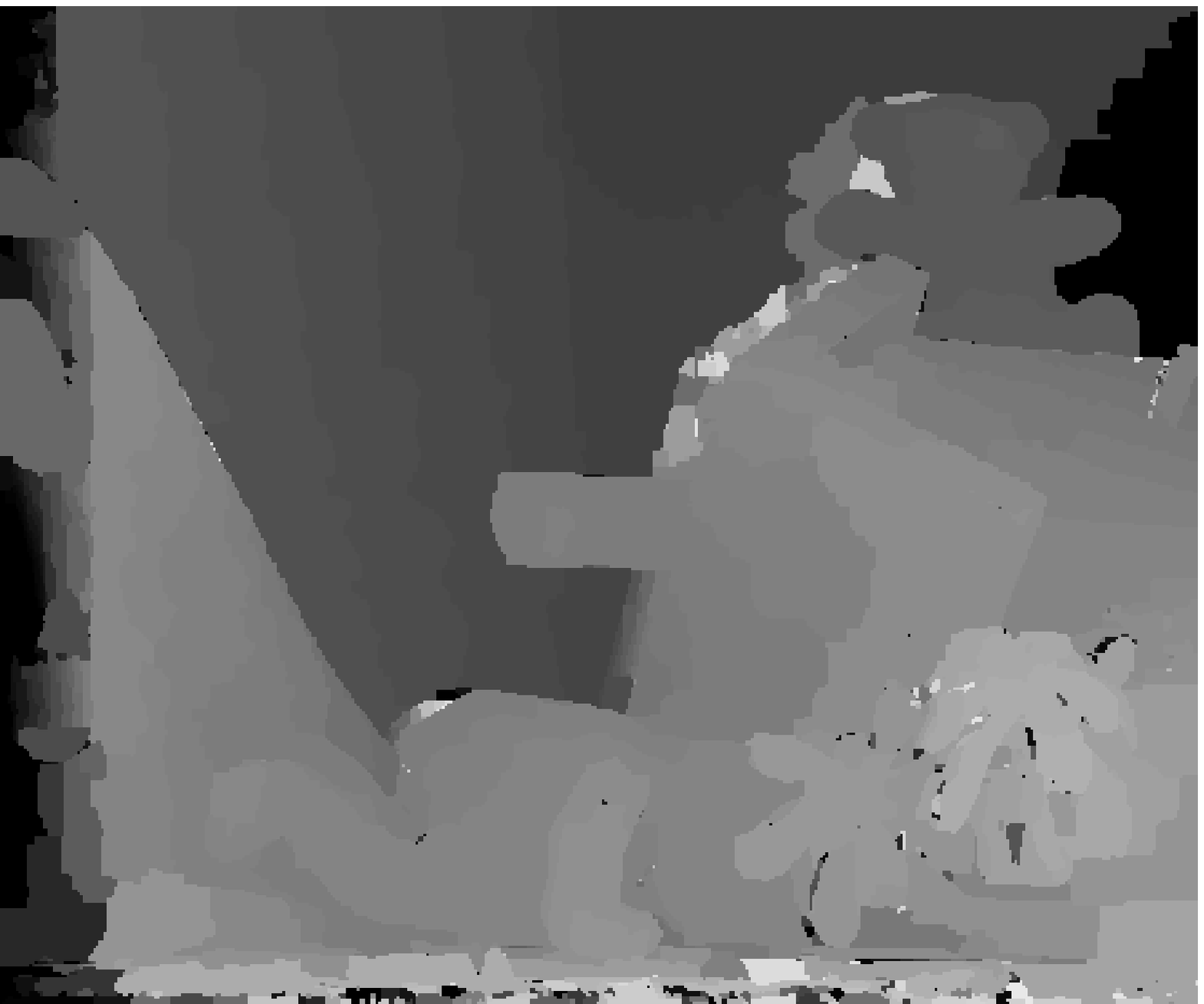}&
\includegraphics[width=\stwidth]{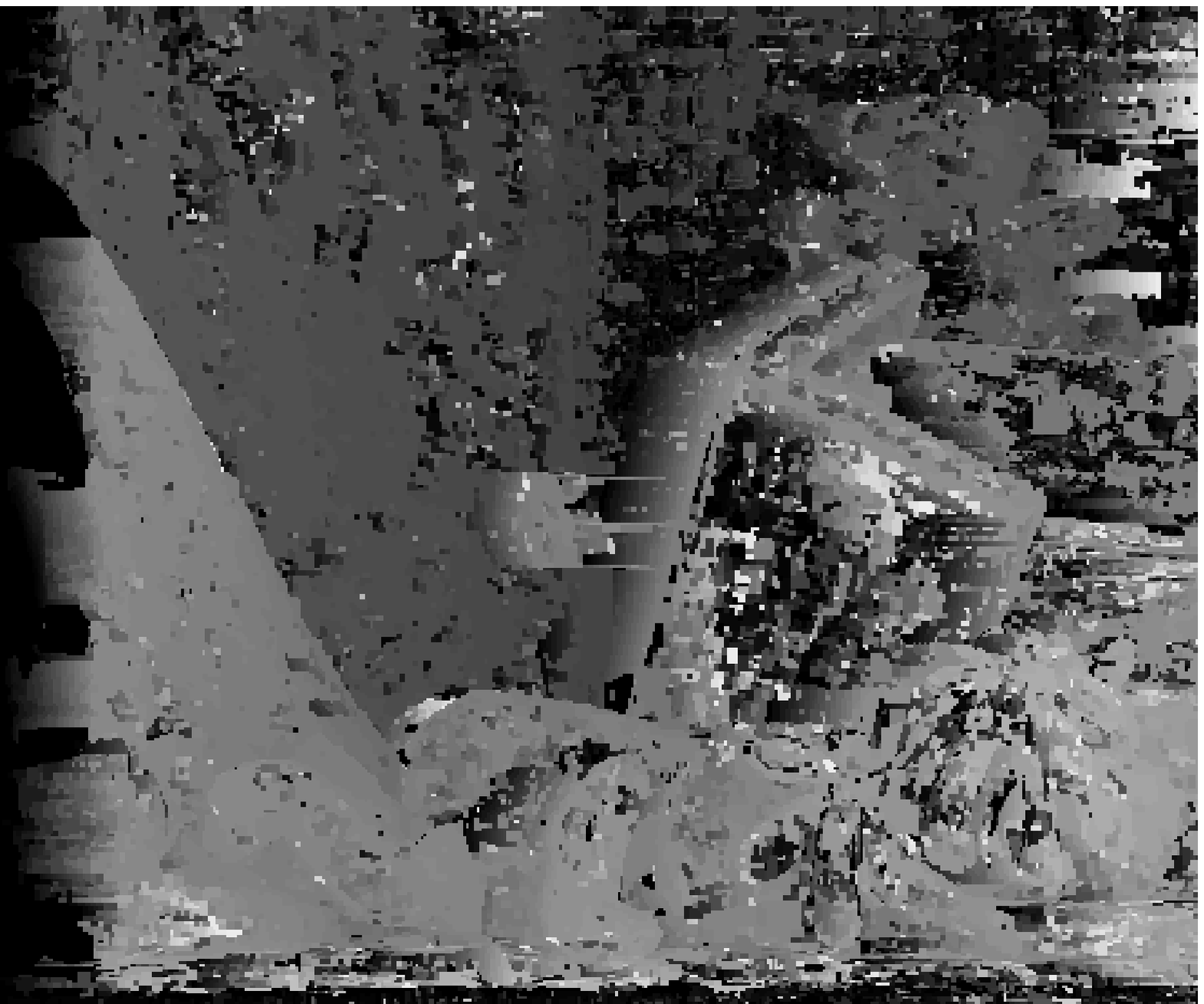}&
\includegraphics[width=\stwidth]{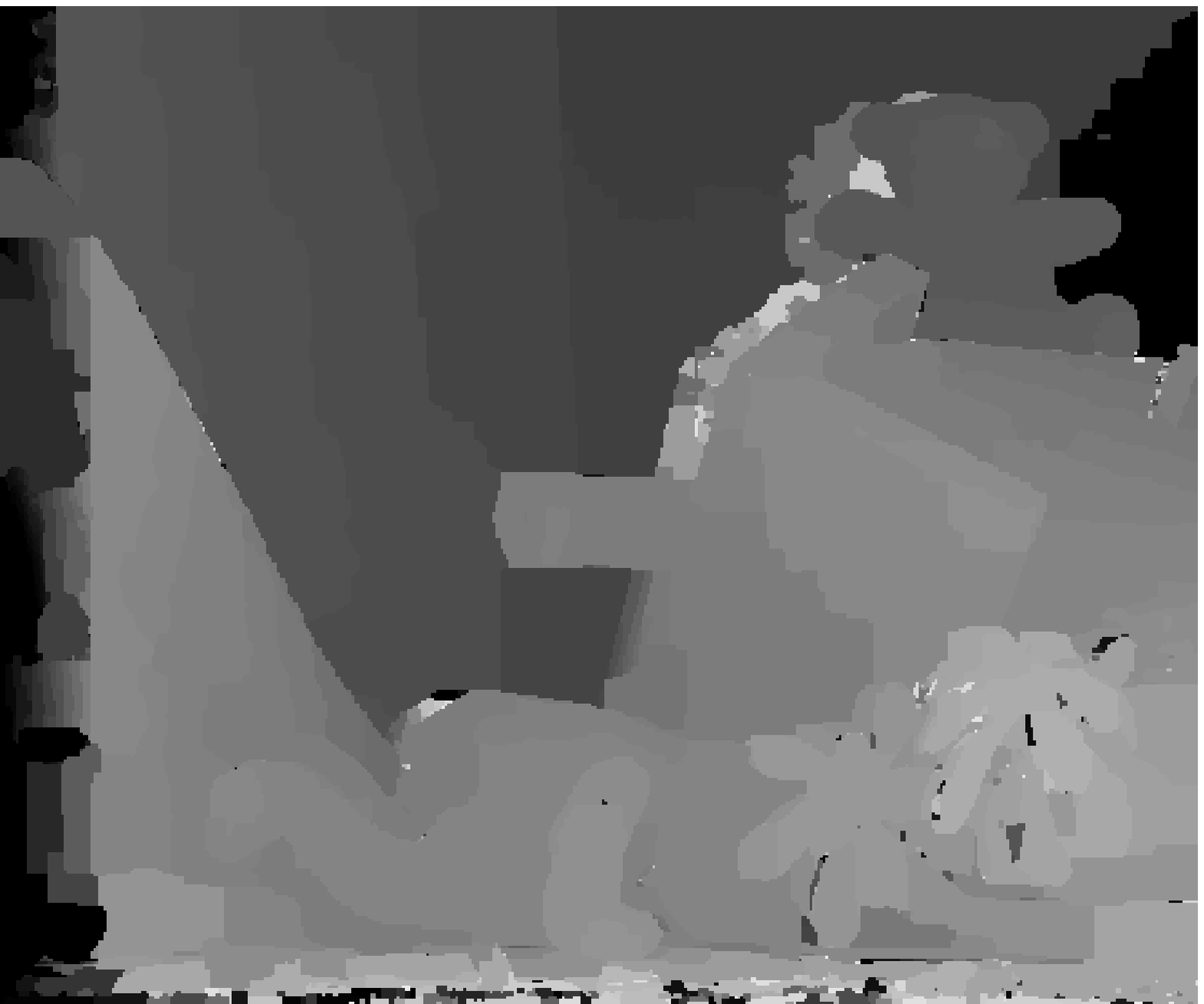}&
\includegraphics[width=\stwidth]{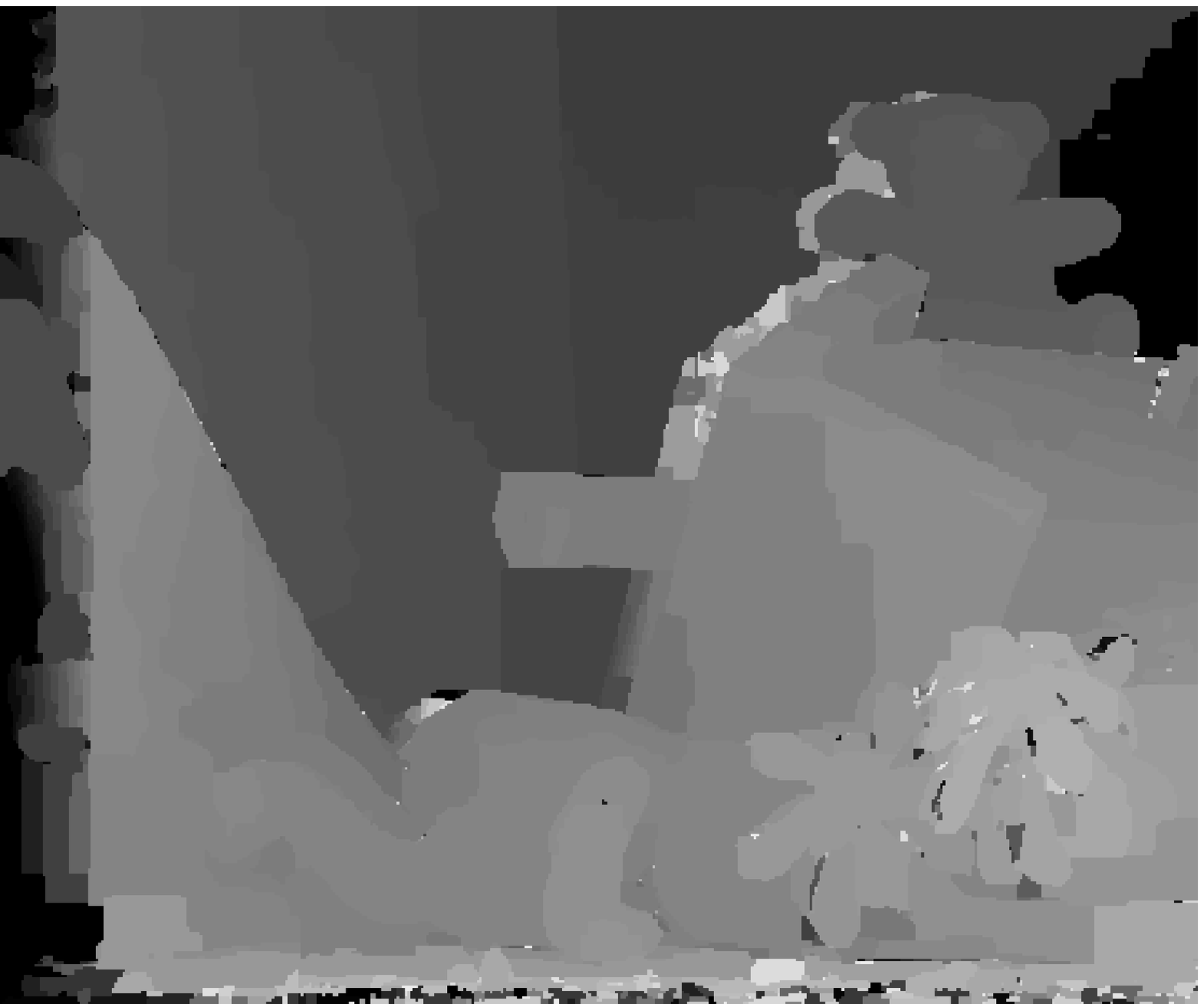}&
\includegraphics[width=\stwidth]{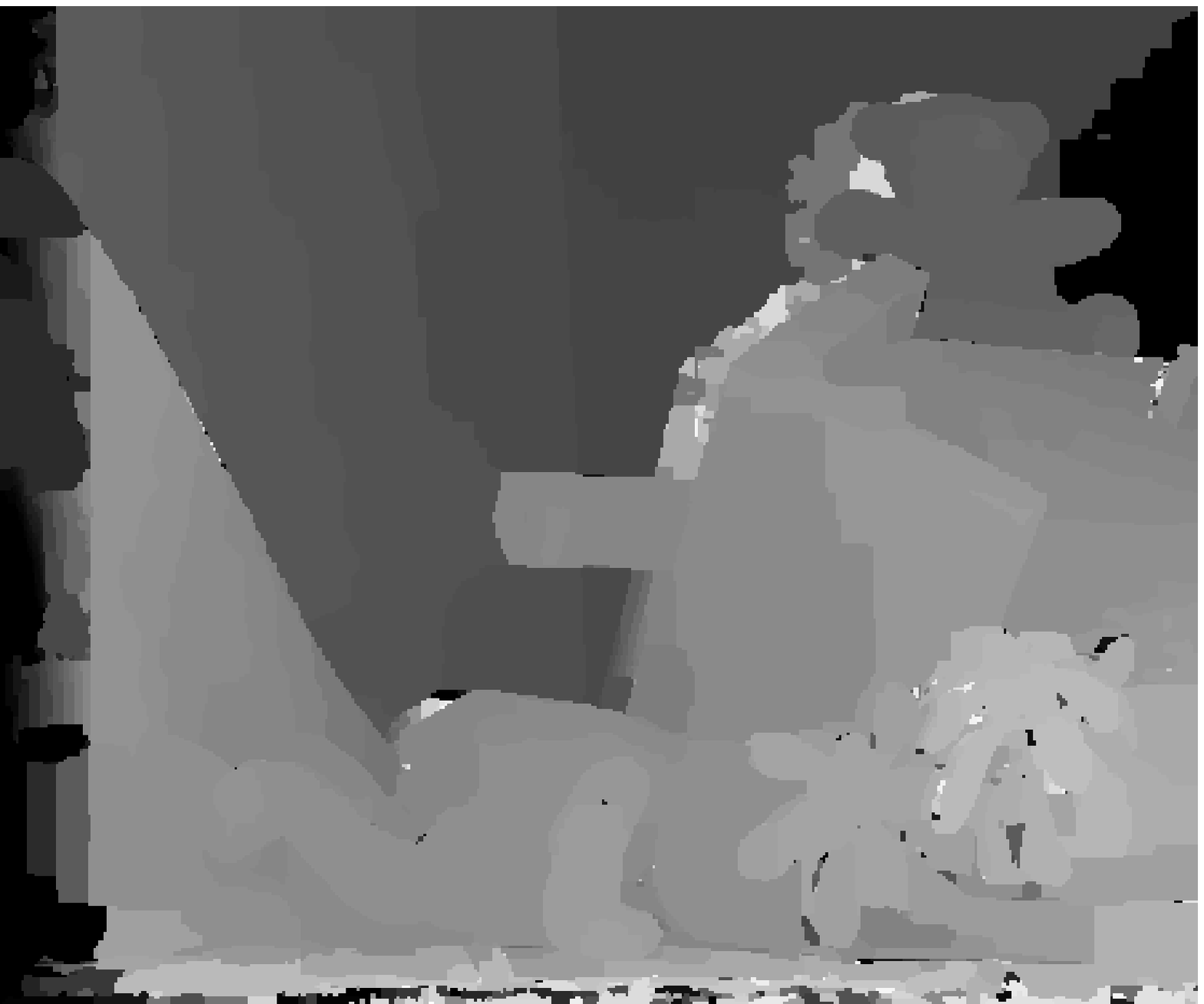}&
\includegraphics[width=\stwidth]{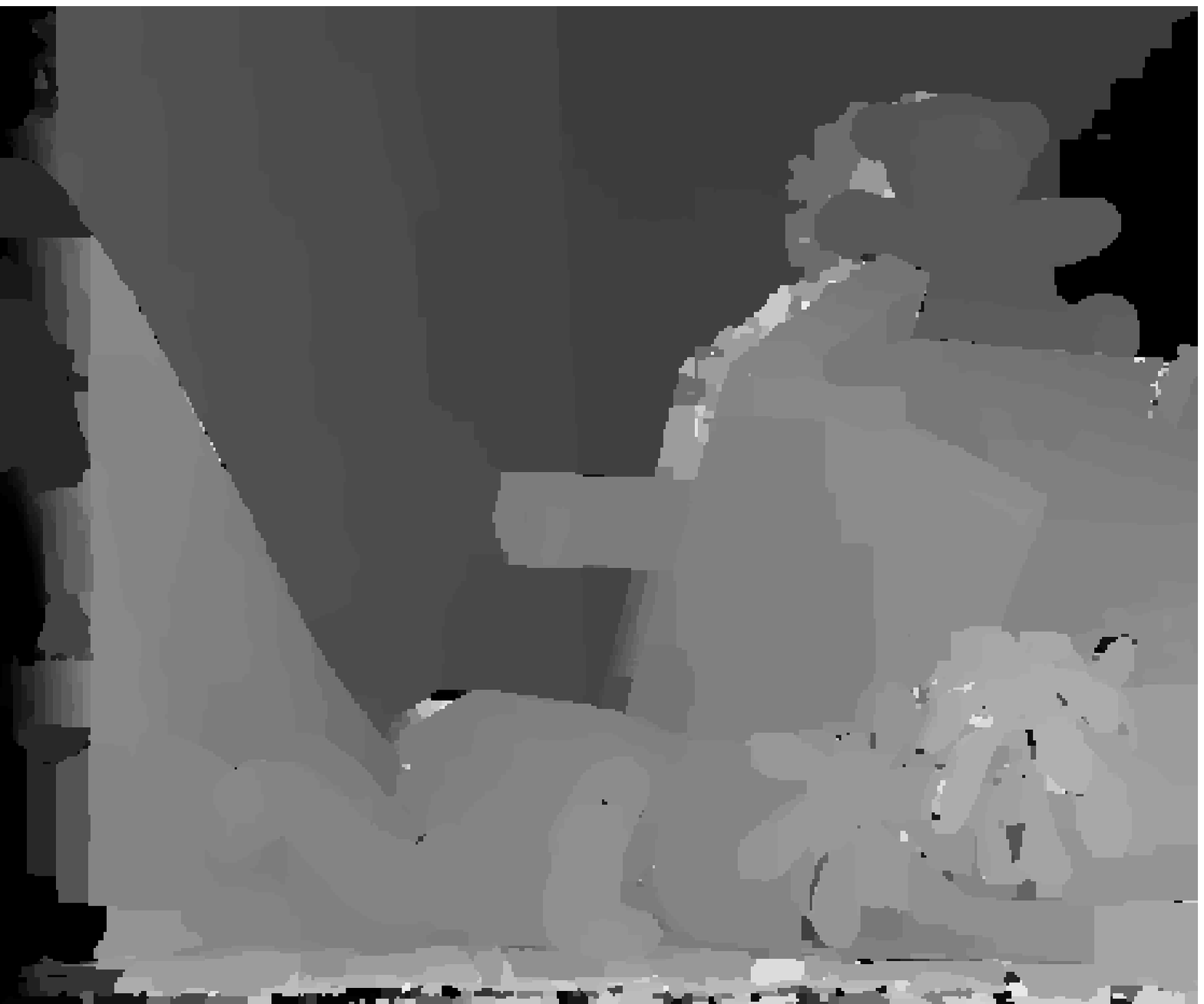}&
\includegraphics[width=\stwidth]{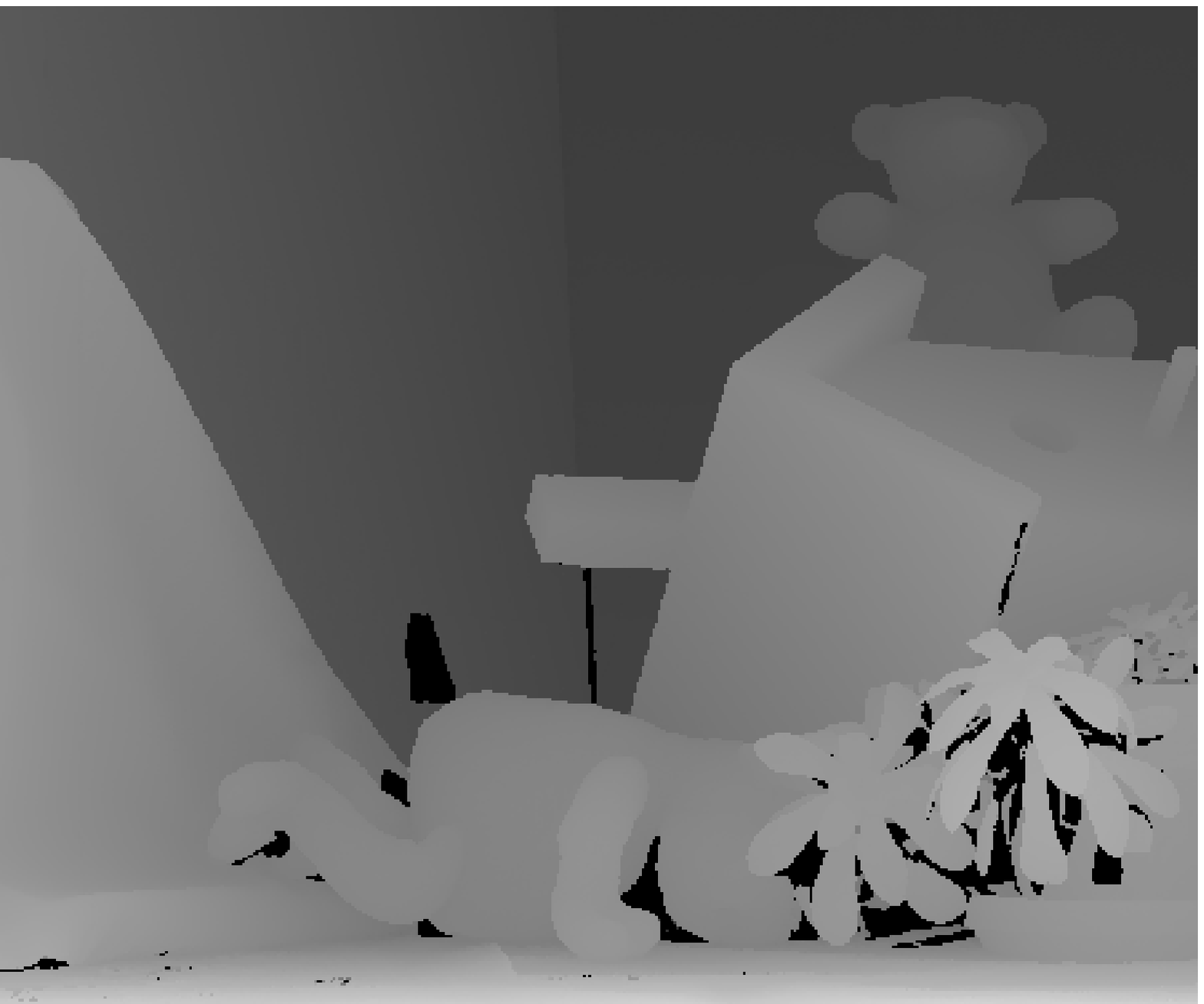}\\
\end{tabular}
\caption{ {\bf Stereo:}
{\em Note how our multiscale framework drastically improves ICM results.
visible improvement for $\alpha\beta$-swap can also be seen in the middle row (Venus).
Numerical results for these examples are shown in Table~\ref{tab:stereo-res}.
Energies from \protect\cite{Szeliski2008}.}}
\label{fig:res-stereo}
\end{figure*}

\begin{figure*}
\centering
\newlength{\dnwidth}
\setlength{\dnwidth}{.12\linewidth}
\begin{tabular}{c||c@{ }c||c@{ }c||c@{ }c}
Input & \multicolumn{2}{c||}{ICM} & \multicolumn{2}{c}{Swap} & \multicolumn{2}{c}{Expand}\\
 & {\color{ours}Ours} & Single scale & {\color{ours}Ours} & Single scale & {\color{ours}Ours} & Single scale\\ \hline
\includegraphics[width=\dnwidth]{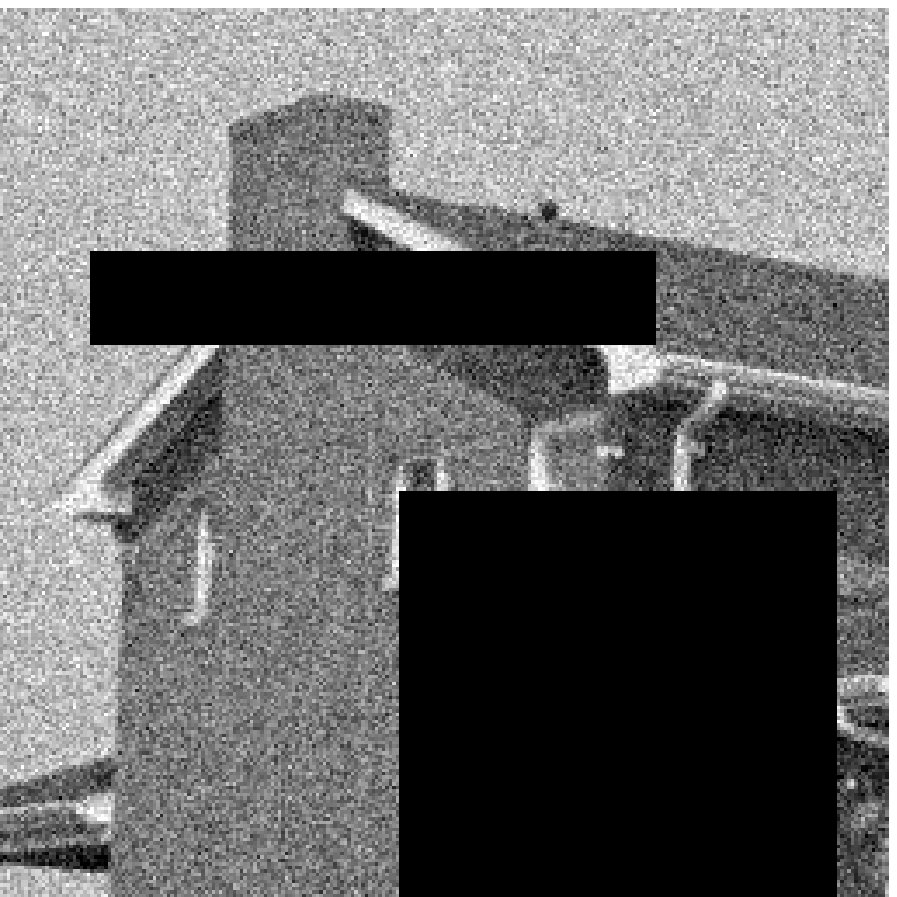}&
\includegraphics[width=\dnwidth]{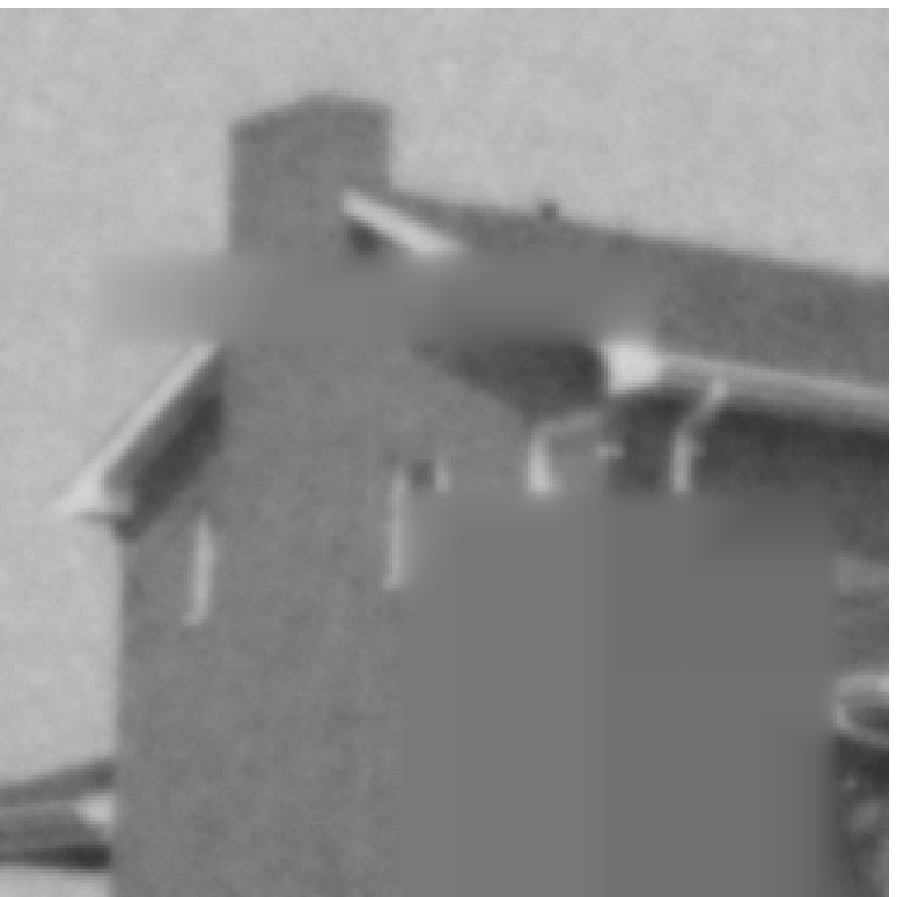}&
\includegraphics[width=\dnwidth]{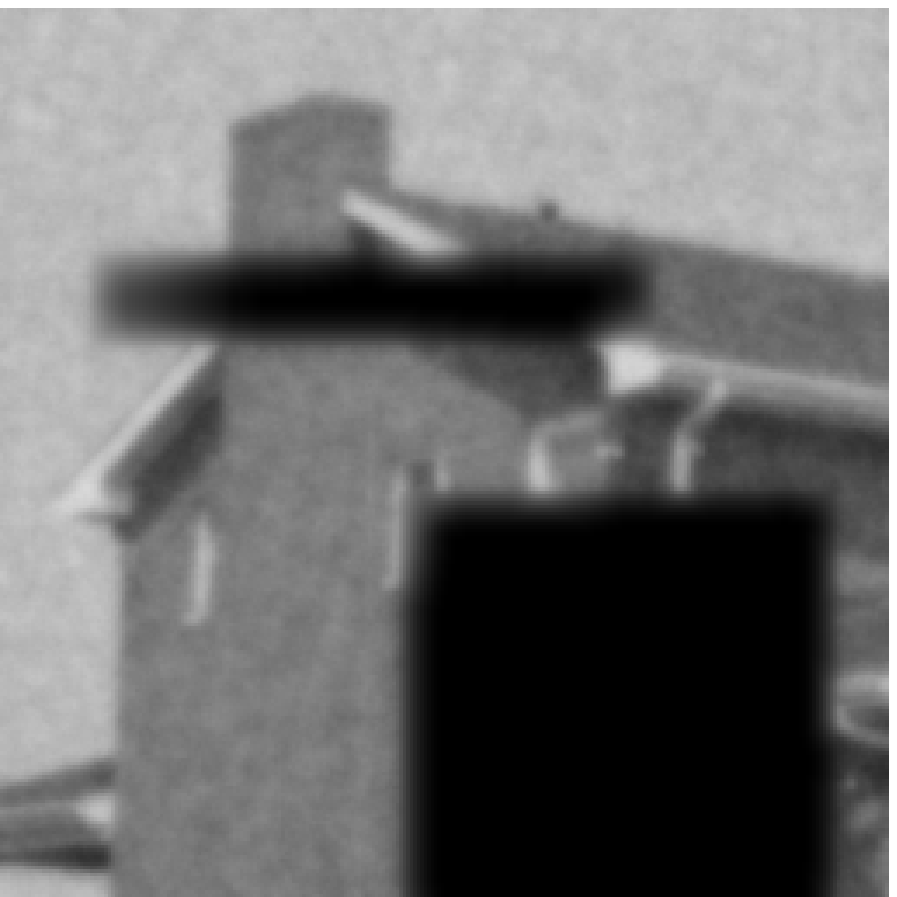}&
\includegraphics[width=\dnwidth]{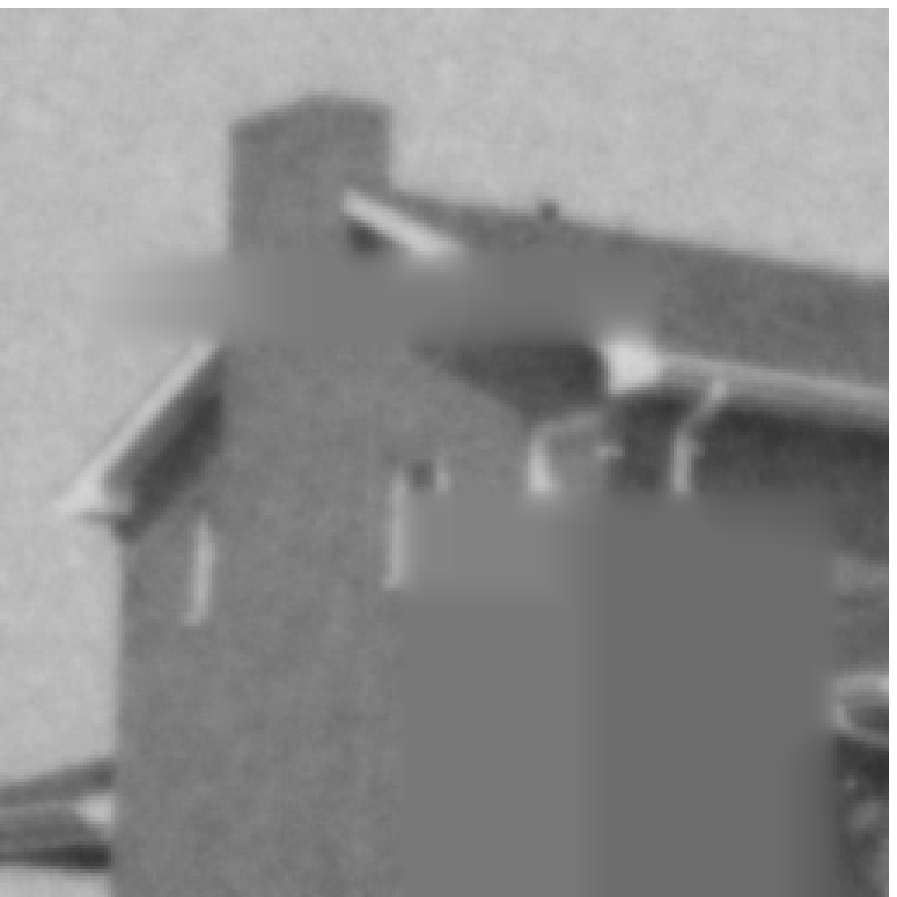}&
\includegraphics[width=\dnwidth]{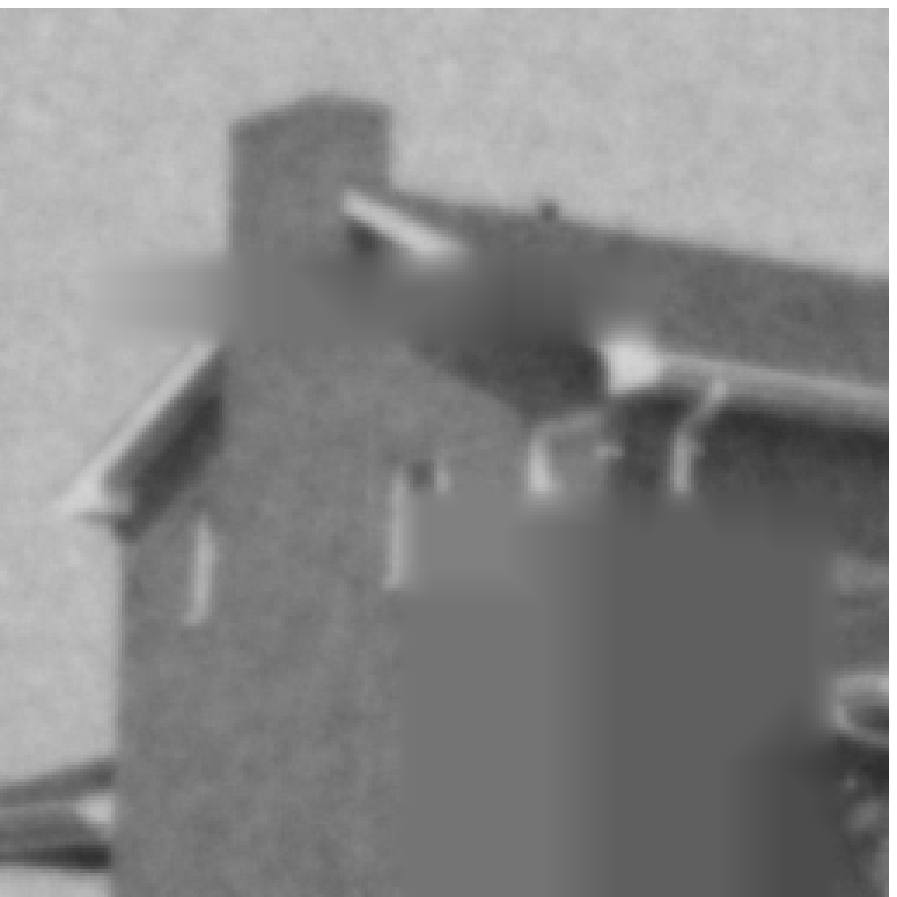}&
\includegraphics[width=\dnwidth]{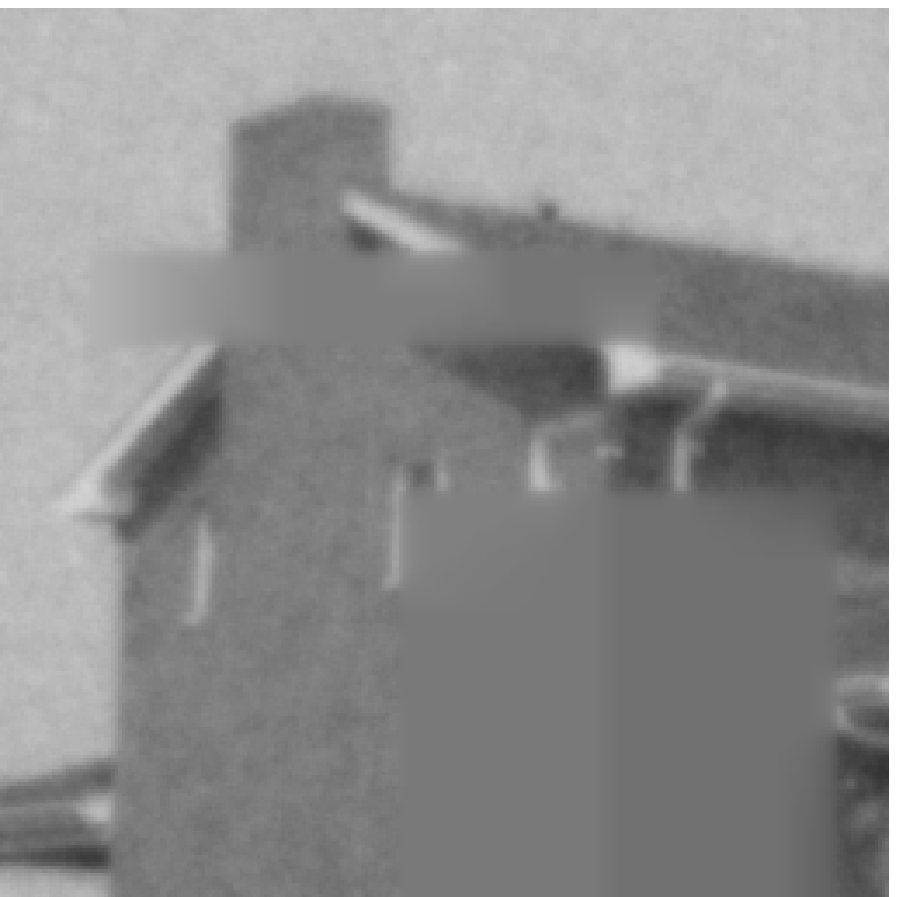}&
\includegraphics[width=\dnwidth]{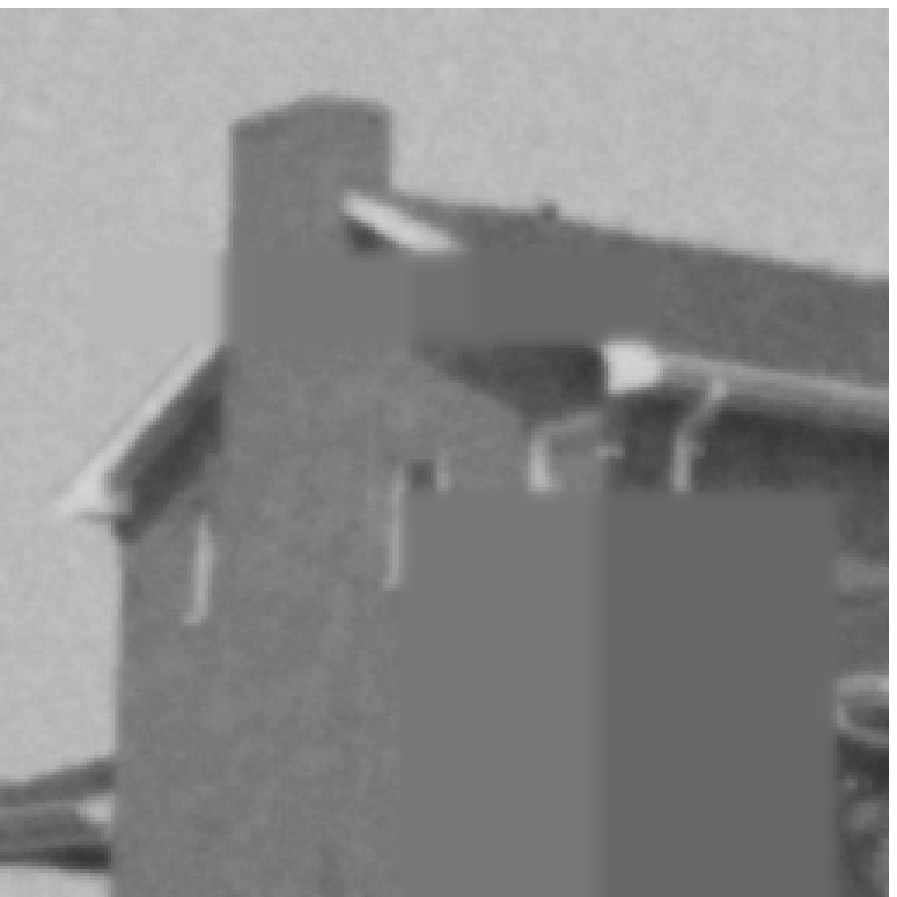}\\
\includegraphics[width=\dnwidth]{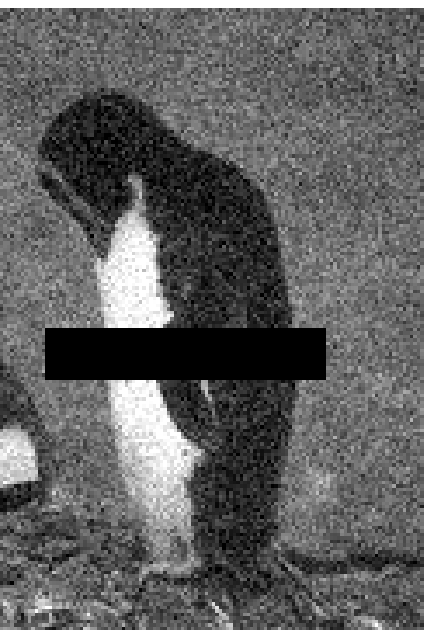}&
\includegraphics[width=\dnwidth]{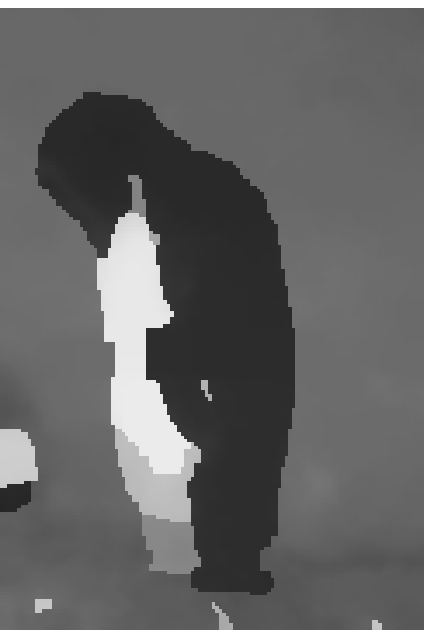}&
\includegraphics[width=\dnwidth]{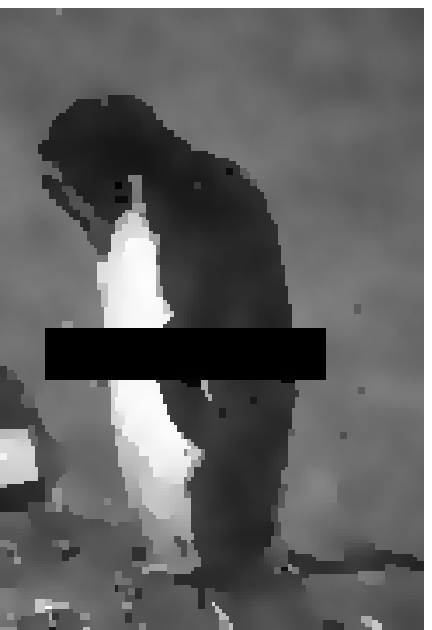}&
\includegraphics[width=\dnwidth]{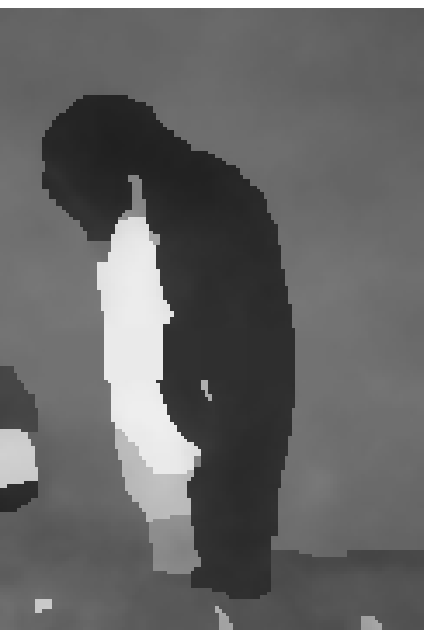}&
\includegraphics[width=\dnwidth]{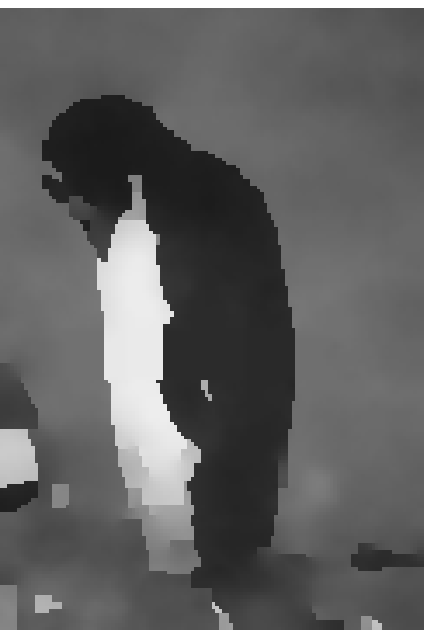}&
\includegraphics[width=\dnwidth]{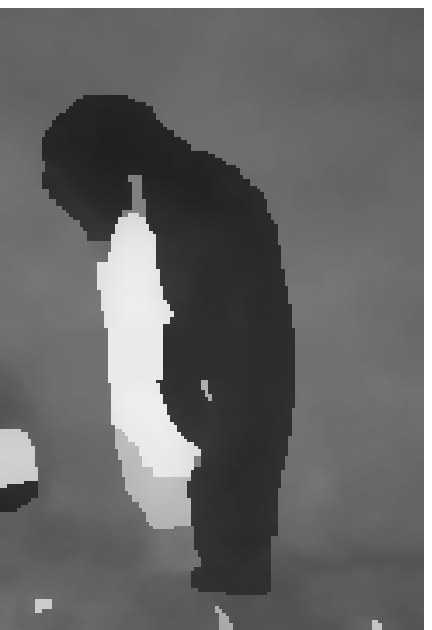}&
\includegraphics[width=\dnwidth]{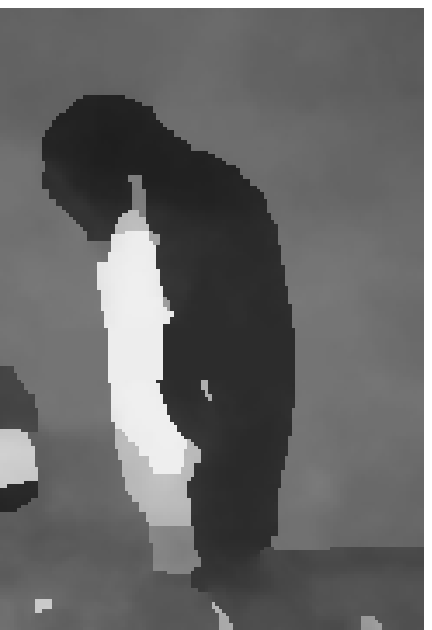}\\
\end{tabular}
\caption{ {\bf Denoising and inpainting:}
{\em
Single scale ICM is unable to cope with inpainting: performing local steps it is unable to propagate information far enough to fill the missing regions in the images.
On the other hand, our multiscale framework allows ICM to perform large steps at coarse scales and successfully fill the gaps.
Numerical results for these examples are shown in Table~\ref{tab:denoise-res}.
Energies from \protect\cite{Szeliski2008}.}}
\label{fig:res-denoise}
\end{figure*}

For these submodular energies, TRW-S (single scale) performs quite well and in fact, if enough iterations are allowed
its lower bound converges to the global optimum.
As opposed to TRW-S, large move making and ICM do not always converge to the global optimum.
Yet, we are able to show a significant improvement for primal optimization algorithms when used within our multiscale framework.
Tables~\ref{tab:stereo-res} and~\ref{tab:denoise-res}
and
Figs.~\ref{fig:res-stereo} and~\ref{fig:res-denoise}
show our multiscale results for the different submodular energies.


\begin{table}
\caption{ {\bf Stereo:}
{\em
Showing percent of achieved energy value relative to the lower bound (closer to $100\%$ is better).
Visual results for these experiments are in Fig.~\ref{fig:res-stereo}.
Energies from \cite{Szeliski2008}.}}
\centering
\setlength{\tabcolsep}{.5mm}
\begin{tabular}{c||c|c||c|c||c|c}
 & \multicolumn{2}{c||}{ICM} & \multicolumn{2}{c||}{Swap}& \multicolumn{2}{c}{Expand}\\
               & {\color{ours}Ours}      & single scale & {\color{ours}Ours}   & single scale & {\color{ours}Ours}   & single scale\\
\hline \hline
Tsukuba & {\color{ours}$102.8\%$} &$653.4\%$ &{\color{ours}$100.2\%$} &$100.5\%$ &{\color{ours}$100.1\%$} &$100.3\%$ \\ \hline
Venus & {\color{ours}$112.3\%$} &$405.1\%$ &{\color{ours}$102.8\%$} &$128.7\%$ &{\color{ours}$102.7\%$} &$102.8\%$ \\ \hline
Teddy & {\color{ours}$102.5\%$} &$234.3\%$ &{\color{ours}$100.4\%$} &$100.8\%$ &{\color{ours}$100.3\%$} &$100.5\%$ \\ \hline
\end{tabular}
\label{tab:stereo-res}
\end{table}

\begin{table}
\caption{ {\bf Denoising and inpainting:}
{\em
Showing percent of achieved energy value relative to the lower bound (closer to $100\%$ is better).
Visual results for these experiments are in Fig.~\ref{fig:res-denoise}.
Energies from \protect\cite{Szeliski2008}.} }
\centering
\setlength{\tabcolsep}{.5mm}
\begin{tabular}{c||c|c||c|c||c|c}
 & \multicolumn{2}{c||}{ICM} & \multicolumn{2}{c||}{Swap} & \multicolumn{2}{c}{Expand}\\
               & {\color{ours}Ours}      & single scale & {\color{ours}Ours}   & single scale & {\color{ours}Ours}   & single scale\\
\hline \hline
House & {\color{ours}$100.5\%$} &$111.3\%$ &{\color{ours}$100.4\%$} &$100.9\%$ &{\color{ours}$102.3\%$} &$103.4\%$ \\ \hline
Penguin & {\color{ours}$106.9\%$} &$132.9\%$ &{\color{ours}$104.6\%$} &$111.3\%$ &{\color{ours}$104.0\%$} &$103.7\%$ \\ \hline
\end{tabular}
\label{tab:denoise-res}
\end{table}

\subsection{Comparing variable agreement estimation methods}

As explained in Sec.~\ref{sec:matrix-P} the agreements between the variables are the most crucial component in constructing an effective multiscale scheme.
In this experiment we compare our energy\-/aware agreement measure (Sec.~\ref{sec:local-correlations}) to three methods proposed by \cite{Kim2011}: ``unary-diff", ``min-unary-diff" and ``mean-compat".
These methods estimate the agreement based either on the unary term or the pair-wise term, but {\em not both}.
We also compare to an energy-agnostic measure, that is $c_{ij}=1$ $\forall ij\in\EE$,
this method underlies \cite{Felzenszwalb2006,Komodakis2010}.

For each energy we estimate variable agreements using these five different approaches.
These different estimations are then used to construct five different energy-pyramids (as described in Sec.~\ref{sec:amg-p}).
Better agreement estimation will results with better exploration of the multiscale landscape of the energy yielding better optimization results.
We use ICM with each of the five energy-pyramids to evaluate the influence these methods have on the resulting multiscale performance for three representative energies.

\begin{figure}
\includegraphics[width=\linewidth]{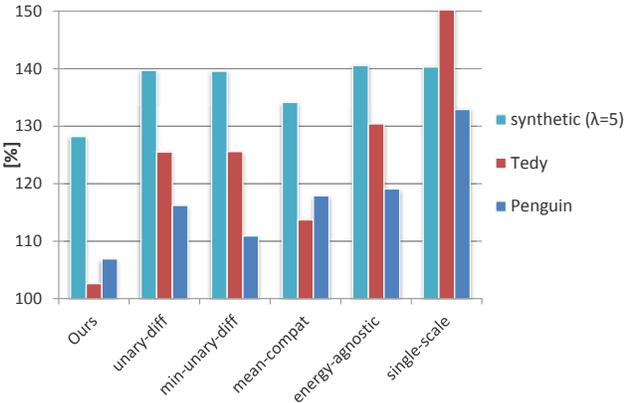}
\caption{{\bf Comparing agreements estimation methods:}
{\em
Graphs showing percent of lower bound (closer to $100\%$ is better)
for different methods of computing variable-agreements.
One bar is cropped at $150\%$.
Our energy\-/aware measure consistently outperforms all other methods.
As a reference, results of single\-/scale optimization are shown on the right.
}}
\label{fig:comapre-weights}
\end{figure}

Fig.~\ref{fig:comapre-weights} shows percent of lower bound for the different energies.
Energy-pyramids constructed based on our agreement estimation method consistently outperforms all other methods, and successfully balances between the influence of the unary and the pair-wise terms.

\subsection{Coarsening labels}
$\alpha\beta$-swap
does not scale gracefully with the number of labels.
Coarsening an energy in the labels domain (i.e., same number of variables, fewer labels) proves to significantly improve performance of $\alpha\beta$-swap, as shown in Table~\ref{tab:coarsening-v}.
For these examples constructing the energy pyramid took only milliseconds, due to the ``closed form" formula for estimating label correlations.

Our principled framework for coarsening labels improves $\alpha\beta$-swap performance for these energies.

\begin{table}
\caption{ {\bf Coarsening labels:}
{\em
Working coarse\-/to\-/fine in the labels domain. We use 5 scales
with coarsening rate of $\sim0.7$.
Number of variables is unchanged.
Table shows percent of achieved energy value relative to the lower bound (closer to $100\%$ is better), and running times.
These results were obtained using $\alpha\beta$-swap for optimizing each scale.}}
\centering
\begin{tabular}{c|c|c||c|c}
\multirow{2}{*}{Energy} & \#labels &  \#labels   & \multirow{2}{*}{{\color{ours}Ours}}& single \\
                        & (finest)   &  (coarsest)   &                                   & scale \\
\hline
Penguin     & \multirow{2}{*}{256} & \multirow{2}{*}{67} & {\color{ours}$103.6\%$}   & $111.3\%$ \\
(denoising) &                      &                     & {\color{ours}$128$} [sec] & $253$ [sec] \\ \hline
Venus       & \multirow{2}{*}{20}  & \multirow{2}{*}{4} & {\color{ours}$106.0\%$}    & $128.7\%$ \\
(stereo)    &                      &                    & {\color{ours}$100$} [sec]  & $130$ [sec]\\
\end{tabular}
\label{tab:coarsening-v}
\end{table}

\section{Conclusion}

This work presents a unified multiscale framework for discrete energy minimization
that allows for efficient and {\em direct} exploration of the multiscale landscape of the energy.
We propose two paths to expose the multiscale landscape of the energy:
one in which coarser scales involve fewer and coarser {\em variables},
and another in which the coarser levels involve fewer {\em labels}.
We also propose adaptive methods for energy\-/aware interpolation between the scales.
Our multiscale framework {\em significantly improves optimization results for challenging energies}.

Our framework provides the mathematical formulation that ``bridges the gap" and relates multiscale discrete optimization and algebraic multiscale methods used in PDE solvers (e.g., \cite{Brandt1986}).
This connection allows for methods and practices developed for numerical solvers to be applied in multiscale discrete optimization as well.

\begin{acknowledgements}
We would like to thank Maria Zontak and Daniel Glasner for their insightful remarks and discussions.
\end{acknowledgements}

\bibliographystyle{spbasic}      
\bibliography{disc_amg}   

\end{document}